\documentclass{article}

\PassOptionsToPackage{numbers,sort&compress}{natbib}
\usepackage[preprint]{neurips_2026}

\usepackage[utf8]{inputenc} % allow utf-8 input
\usepackage[T1]{fontenc}    % use 8-bit T1 fonts
\usepackage{hyperref}       % hyperlinks
\usepackage{url}            % simple URL typesetting
\usepackage{booktabs}       % professional-quality tables
\usepackage{amsfonts}       % blackboard math symbols
\usepackage{nicefrac}       % compact symbols for 1/2, etc.
\usepackage{microtype}      % microtypography
\usepackage{xcolor}         % colors

\usepackage{graphicx}
\usepackage{amsmath}
\usepackage{enumitem}
\usepackage[most]{tcolorbox}
\usepackage{multirow}
\usepackage{subcaption}
\usepackage{booktabs}
\usepackage[table]{xcolor}

\usepackage{algorithm}
\usepackage{algpseudocode}

\usepackage{listings}
\usepackage{xcolor}
\usepackage{tikz}
\usetikzlibrary{shapes.geometric, arrows.meta, positioning, calc}

\lstdefinestyle{jsonstyle}{
    basicstyle=\ttfamily\scriptsize,
    breaklines=true,
    frame=single,
    backgroundcolor=\color{gray!8},
    rulecolor=\color{gray!50},
    string=[s]{"}{"},
    stringstyle=\color{blue!70!black},
    comment=[l]{//},
    commentstyle=\color{green!50!black}\itshape,
    showstringspaces=false,
    tabsize=2,
    xleftmargin=8pt,
    framexleftmargin=8pt,
    aboveskip=8pt,
    belowskip=8pt,
}

\lstdefinestyle{shellstyle}{
    basicstyle=\ttfamily\scriptsize,
    backgroundcolor=\color{black!5},
    frame=single,
    framerule=0.5pt,
    rulecolor=\color{gray!60},
    breaklines=true,
    xleftmargin=8pt,
    framexleftmargin=8pt,
    aboveskip=8pt,
    belowskip=8pt,
}

\lstdefinestyle{memorystyle}{
    basicstyle=\ttfamily\scriptsize,
    breaklines=true,
    frame=single,
    backgroundcolor=\color{blue!3},
    rulecolor=\color{blue!30},
    showstringspaces=false,
    tabsize=2,
    xleftmargin=8pt,
    framexleftmargin=8pt,
    aboveskip=8pt,
    belowskip=8pt,
}

\title{Soap2Soap: Long Cinematic Video Remaking via Multi-Agent Collaboration}

\author{
  Yiren Song$^{1}$ \quad
  Huilin Zhong$^{1}$ \quad
  Kevin Qinghong Lin$^{2}$ \quad
  Haofan Wang$^{3}$ \quad
  Mike Zheng Shou$^{1}$\thanks{Corresponding author.} \\
  \\
  $^{1}$Show Lab, National University of Singapore \\
  $^{2}$University of Oxford \\
  $^{3}$Lovart AI
}

\begin{document}
\maketitle

\vspace{-8mm}

\begin{figure}[htbp]
\centering
\includegraphics[width=1.0\textwidth]{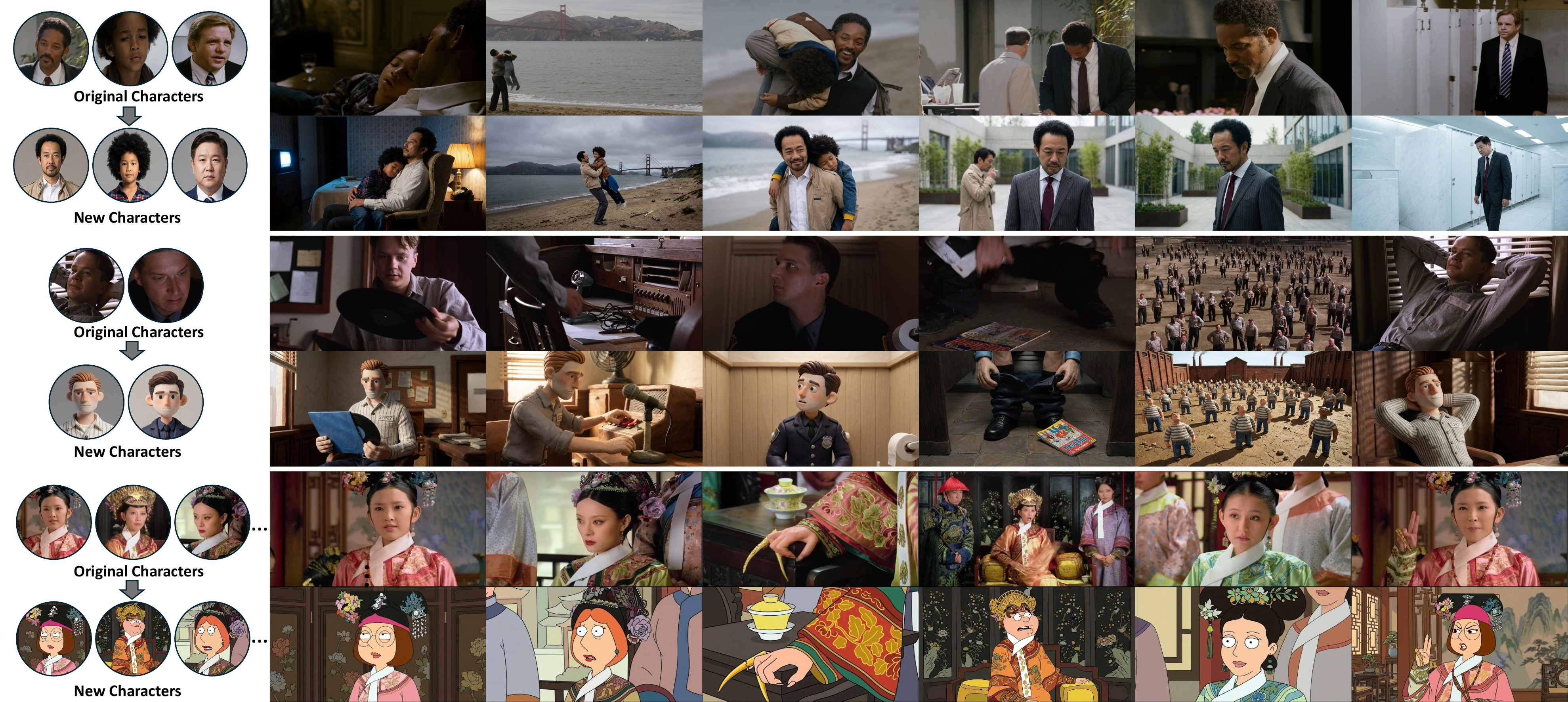}
\vspace{-5mm}
\caption{Soap2Soap supports consistent cinematic remaking across hundreds of shots, maintaining character identity, motion choreography, and narrative fidelity.}
\label{fig:teaser}
\end{figure}

\begin{abstract}
\vspace{-2mm}
We study series-level cinematic remaking, a long-horizon video-to-video generation problem that localizes full episodes or films via stylization or actor replacement while strictly preserving narrative structure, motion choreography, and character identity across hundreds of shots. Existing video generation and editing pipelines often break down in this regime due to compounding identity drift, background mutation, and semantic erosion under large camera motions and viewpoint changes. We propose Soap2Soap, a multi-agent framework that enforces long-term language–visual consistency through a Dual-Bridge Consistency mechanism: a scene-aware JSON screenplay serving as a persistent semantic backbone, and dynamically allocated visual reference anchors at both scene and shot levels. To suppress drift before video synthesis, we introduce batch keyframe consistency, jointly generating multiple keyframes in a shared latent context via a grid-based formulation. A closed-loop verification agent further audits identity, stability, and alignment to trigger selective regeneration. Experiments on SoapBench demonstrate strong improvements over commercial video generation APIs in long-term consistency and narrative fidelity. Code is released at \href{https://github.com/showlab/Soap2Soap}{https://github.com/showlab/Soap2Soap}
\end{abstract}

\section{Introduction}
\vspace{-2mm}

Cinematic remaking transforms an existing film or television series into a localized version through actor replacement, style adaptation, or cultural re-contextualization, while preserving its narrative, choreography, and emotional dynamics. Unlike short video generation, it spans hundreds of shots with complex camera language, multi-character interactions, and long-range narrative dependencies. 

This setting introduces challenges that fundamentally differ from both traditional filmmaking and short-clip generative video models. In real-world production, consistency is largely guaranteed by physical constraints: actors retain stable appearances across shots, and environments remain coherent under continuous lighting and camera setups. In contrast, series-level cinematic remaking must explicitly enforce consistency over extremely long horizons—often spanning hundreds to thousands of shots—where even minor errors can accumulate into severe drift. Concretely, this task requires three coupled forms of long-range consistency: (i) character identity consistency, which requires that each character maintains recognizable visual traits and remains identifiable across drastic viewpoint changes and frequent occlusions; (ii) narrative structure consistency, ensuring that the generated sequence preserves the macro-level temporal logic, causal actions, and semantic continuity of the entire episode; and (iii) motion choreography and camera language consistency, which involves preserving fine-grained movement trajectories and the original cinematic intent, including specific camera angles and transitions. Without explicit mechanisms to maintain these factors jointly, generative remaking systems frequently degrade over time, exhibiting identity mutation, background instability, and semantic erosion.

At a high level, series-level cinematic remaking requires three tightly coupled abilities: (i) \textbf{extremely long-video understanding} at shot granularity, including narrative semantics, character roles, camera intent, and emotional progression; (ii) \textbf{precise character migration} beyond face swapping, where new identities must inherit motion, interactions, and dramatic intent under diverse viewpoints and occlusions; and (iii) \textbf{long-horizon consistency control}, ensuring stable character appearance, environment attributes, and lighting across hundreds of shots. Existing generative pipelines frequently break down once temporal horizons extend beyond a few seconds, exhibiting identity drift, visual mutation, and semantic erosion.

Recent work~\cite{das2023enabling,li2023theory,pan2025agentcoord,tran2025multi,wang2024unleashing} has explored multi-agent systems for long video generation, often mimicking film production workflows by assigning agents~\cite{Lin2023VideoDirectorGPTCM,wu2025automated} roles such as director, screenwriter, or cinematographer. 
While intuitive, this analogy overlooks a key difference: in real filming, consistency is largely guaranteed by physical continuity, whereas in generative remaking, identity, appearance, and scene consistency must be explicitly stored, retrieved, and enforced over long horizons. 
Otherwise, identity drift and appearance mutation quickly accumulate across shots. 
This motivates a first-principles design focused on long-range consistency control. 
Accordingly, we introduce Soap2Soap, a multi-agent framework centered on Dual-Bridge Consistency, which stabilizes generation through a persistent language bridge, i.e., a scene-aware JSON screenplay, and dynamically allocated visual anchors at both scene and shot levels.

Soap2Soap operationalizes the proposed design through three coordinated agents under a shared contextual framework. The Video Understanding Agent analyzes the source video to extract a structured screenplay representation, capturing shot-level narrative events, character roles, and cinematic intent, and assigns shot-specific reference anchors to ensure explicit identity and scene grounding. The Video Generation Agent performs anchor-driven generation in two stages: it first enforces batch keyframe consistency via grid-based joint synthesis to stabilize identity and scene attributes across adjacent shots, and then synthesizes temporally coherent video segments conditioned on the consistent keyframes and contextual memory. Finally, the Verification Agent audits the generated keyframes and video clips against the shared semantic and visual context, and selectively re-generates only the affected shots or local time windows when inconsistencies are detected, forming a closed-loop feedback mechanism that explicitly enforces long-range consistency.

In summary, our contributions are threefold:

\begin{itemize}
    \item We introduce \textbf{long-video remaking}, a movie-scale video-to-video generation task that supports stylization and actor replacement while preserving narrative structure, motion choreography, and character consistency. We further build \textbf{SoapBench} to evaluate long-video understanding and remaking consistency in multi-shot, multi-character scenarios.

    \item We propose \textbf{Soap2Soap}, a multi-agent framework for consistent long-video remaking. It maintains long-horizon visual and narrative coherence through Dual-Bridge Consistency, combining structured JSON screenplays, dynamically allocated visual anchors, contextual memory, and batch keyframe generation.

    \item We conduct extensive experiments and human studies on SoapBench, showing that Soap2Soap outperforms academic baselines and commercial video generation systems in identity stability, scene coherence, and narrative consistency.
\end{itemize}

\section{Related Works}

\subsection{Long Video Understanding}
The paradigm for long video reasoning has shifted from simple frame-level analysis to structured screenplay generation and episodic analysis using Large Video-Language Models (Vid-LLMs)\cite{zhao2023learning,lin2025vlog,chen2023vast,suris2023vipergpt,chen2023video}. While early models relied on handcrafted features, current frameworks like MM-VID\cite{lin2023mm} and ScreenWriter\cite{mahon2024screenwriter} leverage specialized vision-audio tools and Minimum Description Length (MDL)\cite{rissanen1978modeling} principles to segment scenes and identify characters within complex narratives. However, these Vid-LLMs\cite{tang2025video} performing open-ended multi-granularity reasoning often suffer from unidirectional information flow\cite{kalarani2024seeing,min2024morevqa,kim2024image,kahatapitiya2025language} and  LLM-induced hallucinations\cite{zhang2024eventhallusion}. 

\subsection{Long Video Generation}

The development of long-form video generation has progressed through several distinct phases. Early video generation models primarily relied on U-Net diffusion architectures\cite{Blattmann2023StableVD,Guo2023AnimateDiffAY,Wu2022TuneAVideoOT,Ma2025FollowYourClickOR,chen2025transanimate, song2024processpainter}, gradually transitioning to unified generation frameworks like Diffusion Transformers\cite{Wiedemer2025VideoMA,Wang2025WanOA,wu2025hunyuanvideo15technicalreport, song2025worldwander, song2025mitty, yang2025x}. However, computational and modeling limitations typically restricted these models to generating 4–8 second clips. To overcome this constraint, recent work has explored streaming video generation through autoregressive or block-wise-method\cite{huang2025selfforcingbridgingtraintest,Ortega2020DMDAL,gao2025adaworldlearningadaptableworld}, enabling longer sequences but often suffering from error accumulation and inconsistencies. Concurrently, consistency-focused approaches leverage keyframe control, reference guidance\cite{Jiang2025VACEAV,Li2025ICEffectPA,Po2025LongContextSV,yan2026scailstudiogradecharacteranimation,wang2025multishotmaster}, or memory token compression\cite{Fang2025FramePromptIC,Zhang2025FrameCP,zhang2025storymem} to maintain temporal identity and appearance constraints.In this paper, we propose a Video2Video framework for long-duration generation that segments videos into clips while maintaining consistency through explicit memory mechanisms and visual anchors, preserving character identities, scene layouts, and narrative structures over extended sequences.

\subsection{Multi-agent System}
Multi-agent systems~\cite{tran2025multi,das2023enabling,wang2024unleashing,pan2025agentcoord,li2023theory} have recently emerged as a powerful paradigm for decomposing complex generative tasks into modular, role-specialized components. In video generation,Frameworks like VideoDirectorGPT\cite{Lin2023VideoDirectorGPTCM} and MovieAgent\cite{wu2025automated}  established hierarchical production roles and CoT-based planning to enhance narrative logic.To suppress cumulative error propagation, recent works such as AniME\cite{zhang2025animeadaptivemultiagentplanning}, FilmAgent\cite{xu2025filmagentmultiagentframeworkendtoend}, and CoAgent\cite{zeng2025coagentcollaborativeplanningconsistency} introduced global asset memory and closed-loop "plan-execute-verify-refine" collaboration mechanisms. Unlike these text-to-video-focused works, Soap2Soap targets the more demanding "cinematic remaking" task. Our work preserves original motion and narrative through a Dual-Bridge Consistency mechanism—utilizing JSON screenplays and visual anchors—and enforces cross-shot physical consistency via grid-based batch keyframe denoising. This closed-loop framework explicitly manages long-range stability as a systemic objective, ensuring coherence across extensive shot sequences.

\section{Method}

\subsection{Task Definition}

We define \textbf{long-video cinematic remaking} as a long-horizon video-to-video generation task that transforms an existing film or episode into a new version with different actor identities or visual styles, while strictly preserving narrative structure, motion choreography, and audiovisual coherence.

Given a source video $V_{\text{src}}$, the goal is to generate a remade target video $V_{\text{tgt}}$. In this setting, a set of target reference appearances is provided for the main characters. Let $\mathcal{C}_{\text{src}}=\{c_1, c_2, ..., c_K\}$ denote the set of major characters appearing in the source video, and $\mathcal{C}_{\text{tgt}}=\{r_1, r_2, ..., r_K\}$ denote the corresponding set of target reference appearances (e.g., new actor identities or stylized character designs). The objective is to generate $V_{\text{tgt}}$ such that:
(i) the original storyline, shot ordering, camera language, and action causality are strictly preserved;
(ii) each character in $\mathcal{C}_{\text{src}}$ is consistently replaced by its corresponding reference appearance in $\mathcal{C}_{\text{tgt}}$ throughout the entire video; and
(iii) visual appearance, environment, lighting, and audio remain stable over hundreds of shots.

Compared with short-form video generation, long-video remaking introduces additional challenges due to the extreme temporal horizon. Small generation errors can easily accumulate, and this task explicitly evaluates whether a system can maintain semantic fidelity and identity correctness without degrading into identity drift, scene mutation, or narrative corruption.

\begin{figure*}[t]
\centering
\includegraphics[width=1.0\linewidth]{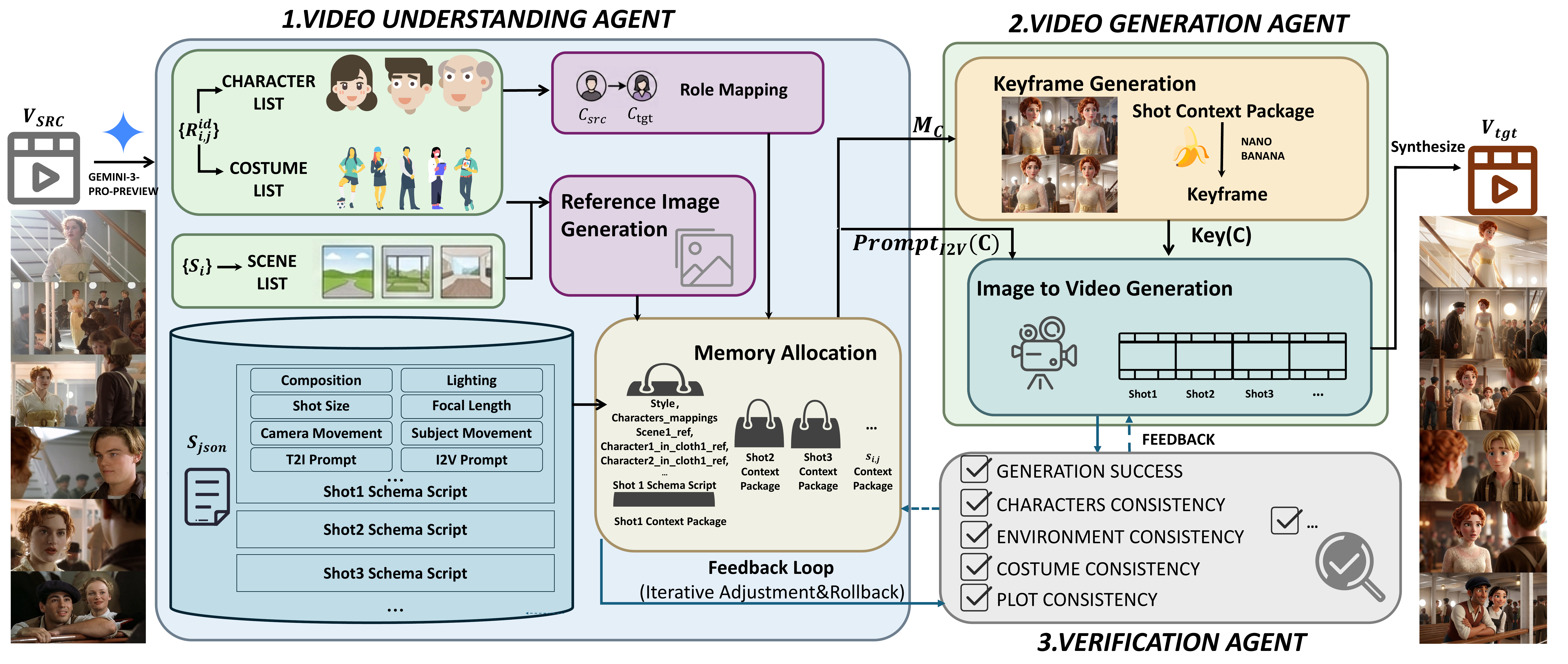}
\caption{
\textbf{Overall architecture of Soap2Soap.}
Soap2Soap decomposes cinematic remaking into three collaborative agents.
The \textbf{Video Understanding Agent} parses long videos into structured textual descriptions and builds multimodal memory for characters, scenes, and context.
The \textbf{Video Generation Agent} uses these memory references to synthesize keyframes and generate video clips via image-to-video generation, which are then stitched into complete sequences.
The \textbf{Verification Agent} evaluates generated keyframes and clips, sending corrective feedback to earlier agents to improve identity consistency, scene stability, and narrative coherence.
}
\vspace*{-1em}
\label{fig_method}
\end{figure*}

\subsection{Overall Architecture}

The overall architecture of Soap2Soap is illustrated in Fig.~\ref{fig_method}. 
Our framework is organized as a collaborative system of three agents: a 
Video Understanding Agent, a Video Generation Agent, and a Verification Agent.

The first two agents form the core pipeline. To enforce long-horizon stability, 
Soap2Soap introduces \textbf{Dual-Bridge Consistency}, which connects video understanding 
and video generation through two shared representations: a structured 
Language Bridge $S_{\text{json}}$ (a scene-aware JSON screenplay) and a 
Visual Bridge $\mathcal{M}$ (a contextual visual memory containing shot-level reference anchors). Based on this shared context, the system performs 
contextual memory allocation, constructing shot-aware memory packages that 
provide the minimal narrative and visual information required for consistent generation.

Conditioned on these anchors and memory units, the Video Generation Agent 
first produces consistent keyframes via anchor-driven synthesis and then generates 
short video segments that are composed into the final remade sequence.

Finally, the Consistency Verification Agent performs closed-loop auditing of 
the generated keyframes and video segments against the shared context 
$(S_{\text{json}}, \mathcal{M})$. When identity drift, scene inconsistency, or semantic 
misalignment is detected, the system selectively regenerates the affected shots, 
ensuring stable long-horizon cinematic remaking.

\subsection{Dual-Bridge Consistency}
We propose \textbf{Dual-Bridge Consistency} as an explicit mechanism to \emph{bridge} the consistency between the input source video and the output remade video over long horizons.
The key idea is to decouple \emph{what should happen} (semantic and cinematic structure) from \emph{how it should look} (visual appearance),
and enforce both through two complementary bridges: a \textbf{Language Bridge} and a \textbf{Visual Bridge}.

\paragraph{Language Bridge.}
We represent the persistent semantic backbone as a scene-aware JSON screenplay $S_{\text{json}}$, obtained via a long-context video understanding model.
$S_{\text{json}}$ follows a structured schema that records shot-level narrative events, character participation, and cinematic intent,
including camera language (e.g., shot scale, viewpoint, motion), scene descriptions, fine-grained actions, and storyline progression.
Importantly, it also provides explicit generation instructions such as per-shot \texttt{T2I} and \texttt{I2V} prompts,
ensuring that the remade video preserves shot order and action causality even under large appearance changes.

\paragraph{Visual Bridge.}
To prevent per-shot re-sampling of identity and scene attributes, we maintain an external visual memory $\mathcal{M}$ with explicit anchors.
At the \emph{scene level}, $\mathcal{M}$ stores environment reference images that stabilize background layout, lighting, and overall style.
At the \emph{shot level}, it assigns character reference images for all appearing identities, together with keyframes as additional visual anchors.
We denote the anchor set for shot $s_{i,j}$ as $\mathcal{M}_{i,j}$, which provides the minimum necessary visual constraints for consistent generation. For each shot $s_{i,j}$, Soap2Soap enforces the invariant that generation must be conditioned on both the semantic specification
$S_{\text{json}}(s_{i,j})$ and the allocated visual anchors $\mathcal{M}_{i,j}$.
This dual-bridge design turns long-range consistency from an emergent by-product of prompting into an explicit, controllable objective.

\subsection{Contextual Memory Allocation} Long videos typically involve multiple characters, scenes, and narrative threads. Directly loading the entire global context for every shot is both redundant and unstable, often leading to role confusion and degraded visual generation quality. In practice, different shots require only a subset of the global narrative and visual information. To address this issue, the Video Understanding Agent includes a \textbf{context-aware memory allocation mechanism} that dynamically constructs compact memory packages for each shot. Instead of using a global reference set, the system selectively retrieves the minimal yet sufficient context required for generating the current shot. Formally, for a shot $s_{i,j}$, the allocated memory package is defined as 
{\setlength{\abovedisplayskip}{2pt}
\setlength{\belowdisplayskip}{2pt}
\setlength{\abovedisplayshortskip}{2pt}
\setlength{\belowdisplayshortskip}{2pt}
\begin{equation}
\mathcal{M}_{i,j} = \texttt{Allocate}(S_{\text{json}}, s_{i,j}),
\end{equation}
}
where the allocation module analyzes the narrative context in $S_{\text{json}}$ and determines the relevant information needed for the current generation step. Each memory package $\mathcal{M}_{i,j}$ contains both semantic instructions and visual reference anchors required for generation, including the shot description, the Text-to-Image prompt for keyframe generation, the Image-to-Video prompt describing motion and camera dynamics, identity references for characters appearing in the shot (including character ID and appearance cues), and scene reference images that stabilize background layout, lighting, and overall scene style. By allocating shot-specific memory packages, the system avoids loading irrelevant history while preserving the necessary narrative and visual context, which significantly improves generation stability for multi-character and multi-scene long-video remaking.

\subsection{Anchor-Driven Visual Generation}

Soap2Soap performs \textbf{anchor-driven} visual rendering in two stages: keyframe generation and shot-level video synthesis. Both stages strictly condition on the allocated contextual memory $\mathcal{M}_{i,j}$ to suppress identity drift and scene mutation under frequent shot transitions.

\paragraph{Keyframe generation.}
To improve intra-scene consistency under large viewpoint changes (e.g., reverse shots), we adopt a \textbf{grid joint synthesis} strategy. We group $4$ or $9$ frames that share the same scene and characters and generate them as a single $2\times2$ or $3\times3$ grid in one pass. This design effectively produces a high-resolution keyframe canvas at once, while allowing all sub-images to \emph{share attention} within the same generation context, leading to stronger identity and scene consistency across the grid. The generated grid is then split into individual keyframes $I_K$ for downstream synthesis.

\paragraph{Shot-level video synthesis.}
Given the generated keyframe for shot $s_{i,j}$, we perform image-to-video (I2V) synthesis to produce a 4--8 second clip $V_{i,j}$. Concretely, we call Veo 3 to generate each shot independently and then stitch all shot clips in temporal order to obtain the final remade video. During I2V generation, Veo 3 supports multiple reference images; we therefore condition the model on the shot-level memory $\mathcal{M}_{i,j}$, including the scene/character reference anchors, together with the keyframe as the primary visual guide. The I2V prompt $P_{\text{I2V}}$ is derived from the video understanding model via $S_{\text{json}}$, which specifies camera motion, character actions, and shot dynamics to faithfully reproduce the original choreography.

\subsection{Verification Agent}

To prevent generation failures from compounding, the Verification Agent orchestrates a closed-loop Critique--Correct--Verify mechanism. For each shot $s_{i,j}$, it evaluates the generated keyframes $I_K$ and video $V_{i,j}$ against the context $(S_{\text{json}}, \mathcal{M}_{i,j})$ across four dimensions:
(i) Generation Quality: verifying artifact-free rendering and correct character counts via $S_{\text{json}}$;
(ii) Identity \& Appearance: matching face IDs and clothing to visual anchors $R^{\text{id}}_{i,j}$;
(iii) Environmental \& Style Stability: ensuring background and style consistency with scene anchors; and
(iv) Plot Consistency: confirming actions and interactions faithfully reflect $S_{\text{json}}$.

Upon detecting inconsistencies (e.g., identity drift or plot deviation), the agent formulates structured textual feedback $\Delta$. This feedback is routed to the Understanding and Generation agents to refine the screenplay and spatial controls, respectively. The conditioning context is updated as:
{\setlength{\abovedisplayskip}{3pt}
\setlength{\belowdisplayskip}{3pt}
\begin{equation}
(\mathcal{M}_{i,j}', P_{\text{I2V}}') =
\text{Refine}(\mathcal{M}_{i,j}, S_{\text{json}}, \Delta).
\end{equation}
}
% \begin{equation}
% (\mathcal{M}_{i,j}', P_{\text{I2V}}') = \text{Refine}(\mathcal{M}_{i,j}, S_{\text{json}}, \Delta).
% \end{equation}
The Video Generation Agent then performs  regeneration for the affected shot $s_{i,j}$. The maximum number of retries can be configured to balance generation quality and computational cost. This iterative loop continues until all criteria are met, enforcing long-term consistency without full sequence rollbacks.

\section{SoapBench: A Benchmark for Long-Video Remaking}

Since Soap2Soap operates as a training-free framework, to systematically evaluate long-form cinematic remaking, we introduce SoapBench, an evaluation benchmark designed for long-video remaking under complex multi-shot narratives. Successful long-video remaking requires understanding long-form video content, including shot boundaries, scenes, character identities, narrative structure, and camera motion. Therefore, SoapBench includes two evaluation tracks: Long Video Understanding and Long Video Remaking.

For the understanding track, we collect 10 movies with high-quality scripts from IMDb together with their corresponding detailed screenplays. These scripts provide structured annotations of scenes and characters, enabling shot-level evaluation of character-level understanding. We manually verify the correctness of character annotations to ensure reliable ground truth for the benchmark. Across the movies, this yields a total of 607 shots, with over 95\% of the shots containing explicit character anchors. This facilitates our focus on whether a model can correctly recognize characters and maintain long-range identity consistency across multi-shot video segments.

For the remaking track, considering the cost of large-scale commercial API calls, we use 10 movies and extract one continuous 1.5--5 minute segment from each for remaking (featuring up to 42 consecutive shots in the longest sequence; see Supplementary Fig.~\ref{fig:keyframe_result_6}). We primarily evaluate cross-shot character identity and appearance consistency, scene consistency, and narrative fidelity.

SoapBench covers two representative remaking scenarios:
(1) real-to-stylized transformation, where live-action segments are remade into distinct visual styles (e.g., LEGO or Disney-like rendering); and
(2) live-action re-casting, where target character reference images are provided to generate a realistic ``re-shot'' version while preserving the original motion choreography and narrative structure.

\section{Experiments}

\subsection{Implementation Details}
The Soap2Soap framework is implemented as a heterogeneous multi-agent system using foundation models that represented the state-of-the-art during the time of our study.

For video understanding, we utilize Google Gemini 3 Flash~\cite{google2025gemini3flash}, leveraging its native multimodal architecture, million-token context window, and the cost-efficiency of the Flash variant. We first process the long video to obtain global context and construct a hierarchical memory structure, followed by shot-by-shot analysis to produce shot-level JSON outputs, including dialogues, camera intent, and character actions.

Keyframe generation and video synthesis are handled by Nano Banana 2~\cite{google2026nanobanana2} and Google Veo 3~\cite{google2025veo3}, respectively. We utilize Nano Banana 2 for keyframe generation due to its superior instruction-following performance in complex multi-subject scenarios. Unlike many generative models that struggle with identity cross-contamination in crowded scenes, Nano Banana 2 maintains high stylistic fidelity and precise character-to-action alignment even when multiple individuals are present in a single frame. We provide it with scene and character references to generate $2\times2$ or $3\times3$ grids, followed by super-resolution enhancement. Finally, Google Veo 3 is used for image-to-video synthesis, selected for its robust spatiotemporal consistency in dynamic view synthesis. It transforms the keyframes into 4--8 second clips guided by the JSON screenplay.  

For closed-loop verification, we also instantiate the Verification Agent using Google Gemini 3 Flash. It acts as an autonomous quality controller by executing two multimodal feedback loops: a semantic loop that audits the JSON output for narrative causality, and a visual loop that cross-checks the generated keyframes and video clips against the JSON structural constraints, triggering selective regeneration if mismatches occur. 

\subsection{Baselines}

As long-form video remaking is a relatively new setting with no directly comparable open-source systems, we include Mocha\cite{xu2026mochaendtoendvideocharacterreplacement} (open-source character replacement via reference-based editing) and two strong commercial video-to-video APIs, Kling O1\cite{team2025kling} and Runway Gen-4\cite{runway2025gen4}, using the same reference images for identity/style guidance to best preserve the input motion and structure. Although SeedDance 2.0\cite{bytedance2026seedance2} supports video remaking, comparison was not feasible due to the lack of API access and strict portrait restrictions.

\begin{figure*}[t] \centering \includegraphics[width=1.0\linewidth]{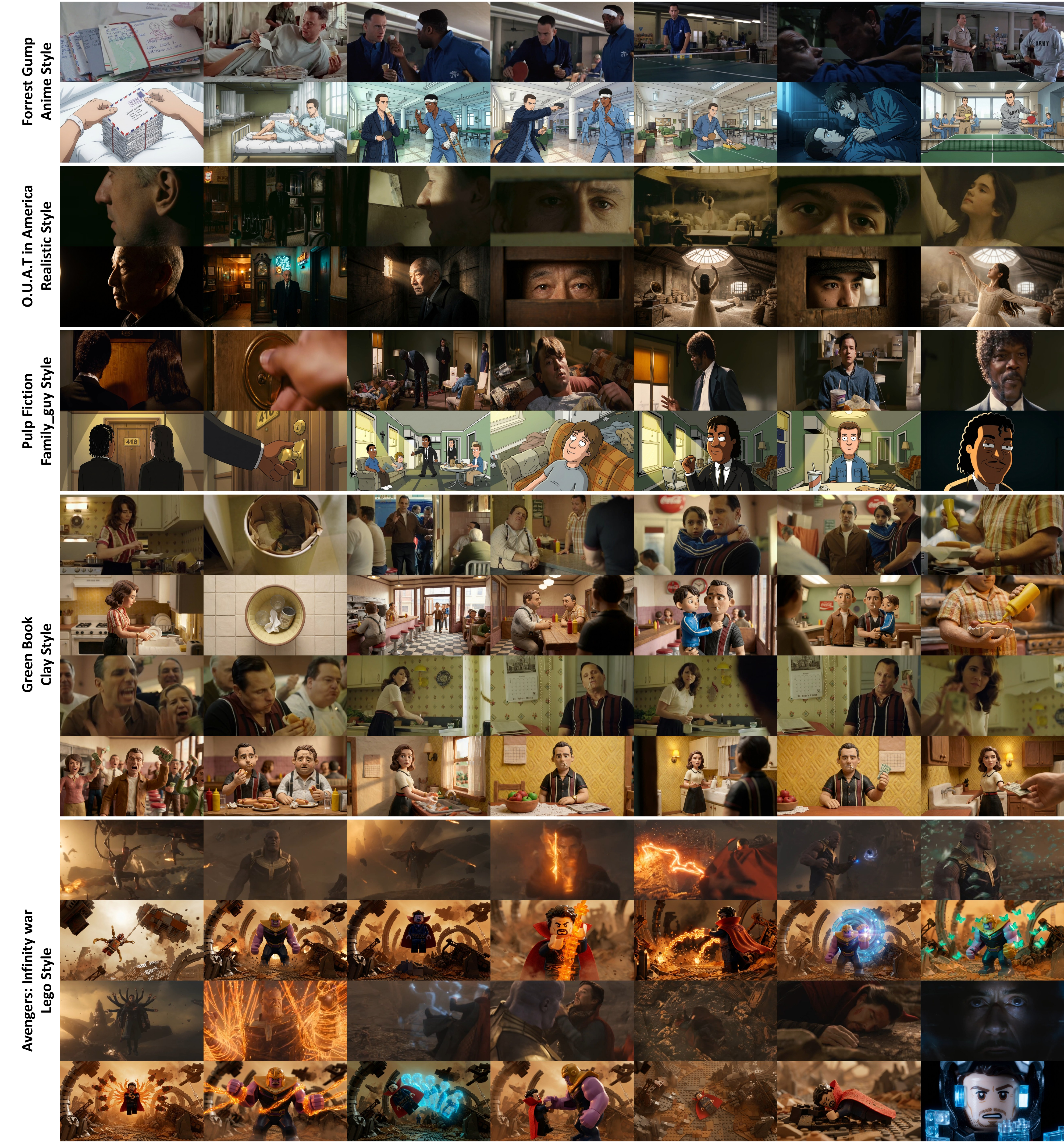} \caption{Results of Soap2Soap on diverse genres.
We demonstrate anime-style remaking, actor swapping,
and multi-character interaction with consistent identity,
scene, and motion preservation.} \end{figure*}

\subsection{Evaluation Metrics}

We evaluate Soap2Soap on SoapBench under two complementary tasks: long-video understanding and long-video remaking. The former measures whether the system can produce correct shot-level structured understanding aligned with the IMDb screenplay (adapted into our shot-indexed script format), while the latter evaluates whether the remade outputs preserve identity and scene consistency under long-horizon generation.

\paragraph{Long-Video Understanding Metrics.} Given each movie, its shot segmentation, and the aligned screenplay, we quantitatively assess understanding quality at the shot level along two axes: character consistency and plot consistency. For character consistency, we focus on the major characters and ignore minor background roles in crowded scenes. We compare the predicted character set of each shot with the ground-truth character list from the script, and report Precision, Recall, IoU, and F1 score. For plot consistency, we evaluate semantic similarity between the predicted storyline descriptions (including the derived T2I/I2V prompts) and the script descriptions using Gemini-based semantic scoring. Together, these metrics reflect whether the system correctly understands who appears and what happens throughout the long video.

\paragraph{Long-Video Remaking Metrics.}
For long-video remaking, we adopt a VLM-as-a-Judge protocol to assess generation quality, using Gemini~3~Flash as the multimodal evaluator. The VLM evaluates identity consistency and scene consistency by jointly analyzing keyframes sampled from the same scene, focusing on cross-shot character stability as well as scene layout and lighting coherence. In addition, we assess plot consistency by examining whether the generated video preserves the overall narrative events and semantic progression of the source sequence. To mitigate potential evaluator bias arising from using the same model family for both generation and evaluation, we further report model-agnostic CLIP Image Score (ID) and CLIP Image Score (Scene) as objective similarity measures. For identity evaluation, we crop detected character regions and compute similarity with the corresponding identity reference images; in multi-character scenes, scores are averaged across all characters. For scene evaluation, we measure similarity between background regions and their corresponding scene anchors. Finally, an extensive User Study is provided in the supplementary material to offer human-aligned cross-validation.

\begin{figure*}[t]
  \centering
  \includegraphics[width=1.0\linewidth]{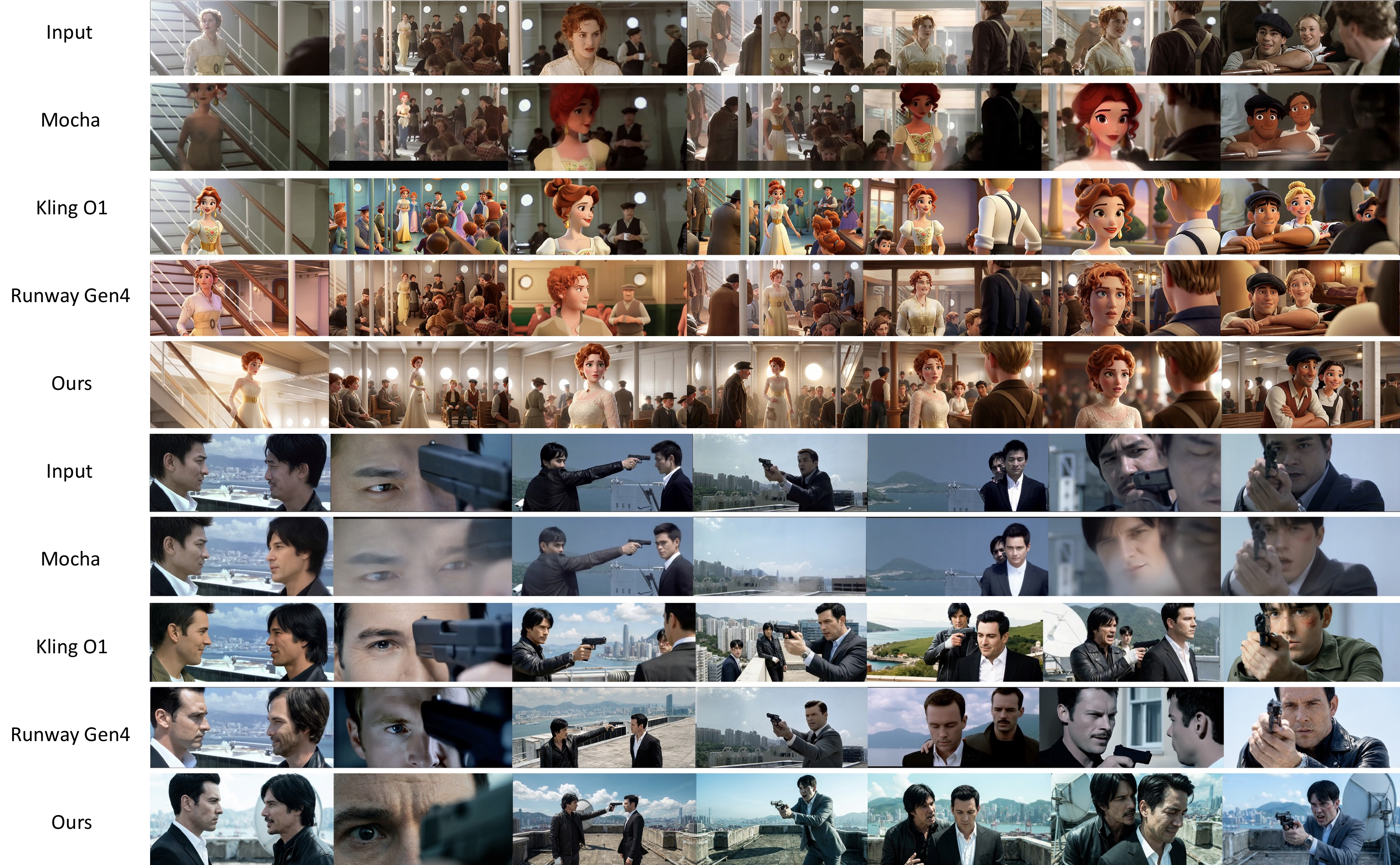}
  \captionof{figure}{Compared with the baseline methods, only our approach achieves accurate cross-shot consistency in long videos, preserving character and environmental visual coherence.}
  \label{fig_compare}
  \vspace{-3mm}
\end{figure*}

\begin{table}[htb]
\centering
\caption{
Quantitative comparison on the long-video remaking task.
We report Gemini-based VLM scores (ID, Scene, Plot) and CLIP Image Scores (CLIP-I)
to evaluate character identity preservation and scene coherence across methods.
VLM scores are rated on a 10-point scale, and the best results are highlighted in bold.
}
\scriptsize
\setlength{\tabcolsep}{3pt}
\renewcommand{\arraystretch}{0.95}

\begin{tabular}{lccccc}
\toprule
Method
& ID-VLM $\uparrow$
& Scene-VLM $\uparrow$
& Plot-VLM $\uparrow$
& CLIP-I (ID) $\uparrow$
& CLIP-I (Scene) $\uparrow$ \\
\midrule
Mocha
& 6.21 & 6.74 & 5.52 & 0.563 & 0.681 \\

Kling O1
& 8.11 & 8.37 & 8.60 & 0.632 & 0.751 \\

Runway Gen4
& 7.48 & 7.96 & 8.27 & 0.604 & 0.724 \\

\midrule
\textbf{Soap2Soap (Ours)}
& \textbf{9.17} & \textbf{8.84} & \textbf{8.67} & \textbf{0.842} & \textbf{0.819} \\
\bottomrule
\end{tabular}
\label{tab:remaking_quant}
\end{table}

\subsection{Quantitative and Qualitative Evaluation}

\paragraph{Quantitative Evaluation.}
Table~\ref{tab:remaking_quant} compares Soap2Soap with Mocha, Kling O1, and Runway Gen4 on the long-video remaking task. 
Our method consistently achieves the best performance across all evaluation metrics. 
In particular, Soap2Soap demonstrates clear advantages in identity preservation and scene consistency, indicating stronger stability of characters and environments across long video sequences. 
It also achieves the highest scores in narrative consistency and visual similarity, showing its ability to maintain coherent story structure and accurate visual appearance during remaking. 
These results validate the effectiveness of our structured language–visual anchoring mechanism for long-horizon video generation.

\paragraph{Qualitative Evaluation.}
Fig.~\ref{fig_compare} presents qualitative comparisons between our method and several strong commercial and open-source baselines. Video editing methods such as Mocha struggle to produce stable actor replacement results, often exhibiting identity instability, artifacts, and unnatural transitions, especially in multi-character scenes. 
State-of-the-art video-to-video models such as Kling O1 and Runway Gen4 generate visually appealing clips, but when long videos are processed in short segments, they frequently suffer from inconsistent backgrounds and identity drift across shots, and stylization becomes unreliable in multi-character scenarios. 
In contrast, Soap2Soap reformulates the task as semantic remaking guided by structured memory and keyframe anchors rather than strict layout-aligned video-to-video generation. 
This design enables more stable character identities, consistent environments, and higher overall visual quality across long sequences.

\begin{figure}[t]
  \centering
  \includegraphics[width=1.0\linewidth]{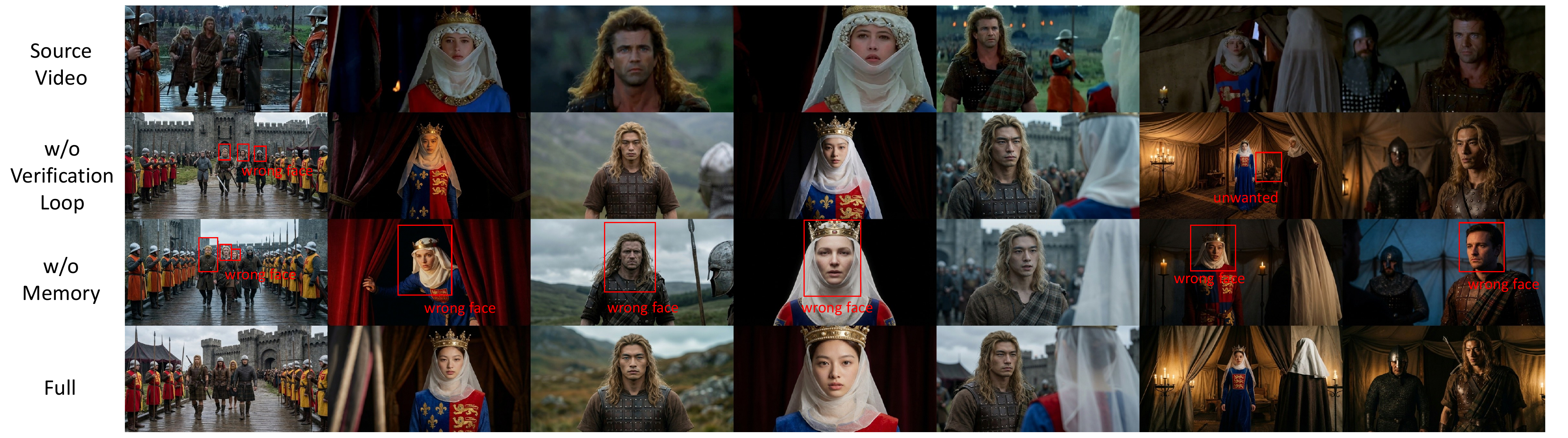}
  \captionof{figure}{Ablation study results, Removing memory allocation causes severe degradation in cross-shot character identity and visual consistency, while removing the verification agent leads to occasional inconsistencies in individual frames.}
  \label{fig_ablation}
\end{figure}

\begin{table*}[htb]
\centering
\caption{
Ablation study on long-video understanding and remaking/generation. 
Understanding metrics are computed at the shot level using IMDb screenplay supervision. 
For remaking/generation, ID-VLM, Scene-VLM, and Plot-VLM denote Gemini 3 Flash-based evaluation scores for identity, scene, and narrative consistency; CLIP-I measures identity and scene similarity. 
VLM scores are rated on a 10-point scale. Best results are highlighted in bold.
}
\scriptsize
\setlength{\tabcolsep}{3pt}
\renewcommand{\arraystretch}{0.95}

\begin{tabular}{lcccc|ccccc}
\toprule
\multicolumn{5}{c|}{\textbf{Long-Video Understanding}} 
& \multicolumn{5}{c}{\textbf{Long-Video Remaking / Generation}} \\
\midrule
\textbf{Setting}
& \textbf{IoU $\uparrow$}
& \textbf{Prec. $\uparrow$}
& \textbf{Recall $\uparrow$}
& \textbf{F1 $\uparrow$}
& \textbf{ID-VLM $\uparrow$}
& \textbf{Scene-VLM $\uparrow$}
& \textbf{Plot-VLM $\uparrow$}
& \textbf{CLIP-I (ID) $\uparrow$}
& \textbf{CLIP-I (Scn) $\uparrow$} \\
\midrule
w/o Dynamic Alloc.
& 0.569 & 0.641 & 0.628 & 0.618
& 5.61 & 4.92 & 7.20 & 0.541 & 0.602 \\

w/o Verification Loop
& 0.874 & 0.893 & 0.894 & 0.887
& 8.91 & 8.53 & 8.32 & 0.826 & 0.801 \\

Full (Soap2Soap)
& \textbf{0.921} & \textbf{0.940} & \textbf{0.943} & \textbf{0.936}
& \textbf{9.17} & \textbf{8.84} & \textbf{8.67} & \textbf{0.842} & \textbf{0.819} \\
\bottomrule
\end{tabular}
\label{tab:ablation_long_video}
\end{table*}

\subsection{Ablation Study}

\paragraph{Video Understanding Agent: Ablating Dynamic Memory Allocation.}
We ablate the Video Understanding Agent by disabling dynamic memory allocation and replacing it with a static global description. As shown in Table~\ref{tab:ablation_long_video}, this non-agentic baseline substantially degrades understanding performance, with IoU dropping from 0.921 to 0.569 and F1 from 0.936 to 0.618. Fig.~\ref{fig_ablation} further shows that removing dynamic allocation leads to identity inconsistencies, indicating that precise shot-level context allocation is essential for maintaining character grounding and narrative continuity in long sequences.

\paragraph{Verification Agent: Ablating the Critique-Correct Loop.}
We further ablate the Verification Agent by removing the critique--correct--verify loop and generating each chunk in a single pass. This leads to consistent degradation in both understanding and generation quality, with F1 decreasing from 0.936 to 0.887 and ID-VLM from 9.17 to 8.91. As illustrated in Fig.~\ref{fig_ablation}, minor identity and semantic errors accumulate over long horizons without verification, demonstrating the importance of closed-loop auditing for robust long-form remaking.

\section{Conclusion}

In this paper, we introduced Soap2Soap, a novel multi-agent framework designed to tackle the long-standing challenges of character identity drift and semantic erosion in cinematic remaking. By decoupling the remaking process into a Dual-Bridge Consistency mechanism that leverages scene-aware JSON screenplays and dynamically allocated visual anchors, our approach ensures both narrative fidelity and visual stability across hundreds of shots. Furthermore, the proposed grid-based batch synthesis effectively mitigates the accumulation of errors inherent in long-sequence generation. Comprehensive evaluations on the SoapBench benchmark demonstrate that Soap2Soap significantly outperforms state-of-the-art academic methods and commercial APIs in maintaining identity and scene coherence. While future work will address fine-grained audio-visual synchronization and extreme motion dynamics, Soap2Soap establishes a scalable and controllable foundation for high-fidelity long-form video transformation.

\newpage
\small
\bibliographystyle{plainnat} 
\bibliography{main}

\begin{thebibliography}{54}
\providecommand{\natexlab}[1]{#1}
\providecommand{\url}[1]{\texttt{#1}}
\expandafter\ifx\csname urlstyle\endcsname\relax
  \providecommand{\doi}[1]{doi: #1}\else
  \providecommand{\doi}{doi: \begingroup \urlstyle{rm}\Url}\fi

\bibitem[Blattmann et~al.(2023)Blattmann, Dockhorn, Kulal, Mendelevitch, Kilian, and Lorenz]{Blattmann2023StableVD}
A.~Blattmann, Tim Dockhorn, Sumith Kulal, Daniel Mendelevitch, Maciej Kilian, and Dominik Lorenz.
\newblock Stable video diffusion: Scaling latent video diffusion models to large datasets.
\newblock \emph{ArXiv}, abs/2311.15127, 2023.
\newblock URL \url{https://api.semanticscholar.org/CorpusID:265312551}.

\bibitem[ByteDance(2026)]{bytedance2026seedance2}
ByteDance.
\newblock Seedance 2.0, 2026.
\newblock URL \url{https://seed.bytedance.com/en/seedance2_0}.
\newblock Accessed: Mar 2026.

\bibitem[Chen et~al.(2023{\natexlab{a}})Chen, Zhu, Haydarov, Li, and Elhoseiny]{chen2023video}
Jun Chen, Deyao Zhu, Kilichbek Haydarov, Xiang Li, and Mohamed Elhoseiny.
\newblock Video chatcaptioner: Towards enriched spatiotemporal descriptions.
\newblock \emph{arXiv preprint arXiv:2304.04227}, 2023{\natexlab{a}}.

\bibitem[Chen et~al.(2023{\natexlab{b}})Chen, Li, Wang, Zhao, Sun, Zhu, and Liu]{chen2023vast}
Sihan Chen, Handong Li, Qunbo Wang, Zijia Zhao, Mingzhen Sun, Xinxin Zhu, and Jing Liu.
\newblock Vast: A vision-audio-subtitle-text omni-modality foundation model and dataset.
\newblock \emph{Advances in Neural Information Processing Systems}, 36:\penalty0 72842--72866, 2023{\natexlab{b}}.

\bibitem[Chen et~al.(2025)Chen, Chen, and Song]{chen2025transanimate}
Xuewei Chen, Zhimin Chen, and Yiren Song.
\newblock Transanimate: Taming layer diffusion to generate rgba video.
\newblock \emph{arXiv preprint arXiv:2503.17934}, 2025.

\bibitem[Das et~al.(2023)Das, Chen, Shyu, and Sadiq]{das2023enabling}
Ayushman Das, Shu-Ching Chen, Mei-Ling Shyu, and Saad Sadiq.
\newblock Enabling synergistic knowledge sharing and reasoning in large language models with collaborative multi-agents.
\newblock In \emph{2023 IEEE 9th International Conference on Collaboration and Internet Computing (CIC)}, pages 92--98. IEEE, 2023.

\bibitem[Fang et~al.(2025)Fang, Gu, and Shou]{Fang2025FramePromptIC}
Guian Fang, Yuchao Gu, and Mike~Zheng Shou.
\newblock Frameprompt: In-context controllable animation with zero structural changes.
\newblock \emph{ArXiv}, abs/2506.17301, 2025.
\newblock URL \url{https://api.semanticscholar.org/CorpusID:280000000}.

\bibitem[Gao et~al.(2025)Gao, Zhou, Du, Zhang, and Gan]{gao2025adaworldlearningadaptableworld}
Shenyuan Gao, Siyuan Zhou, Yilun Du, Jun Zhang, and Chuang Gan.
\newblock Adaworld: Learning adaptable world models with latent actions, 2025.
\newblock URL \url{https://arxiv.org/abs/2503.18938}.

\bibitem[{Google}(2026)]{google2026nanobanana2}
{Google}.
\newblock Gemini 3 flash image (nano banana 2), 2026.
\newblock URL \url{https://ai.google.dev/models/gemini}.
\newblock Developer documentation. Accessed: Mar 2026.

\bibitem[{Google DeepMind}(2025{\natexlab{a}})]{google2025gemini3flash}
{Google DeepMind}.
\newblock Gemini 3 flash: Frontier intelligence built for speed, 2025{\natexlab{a}}.
\newblock URL \url{https://deepmind.google/models/gemini/flash/}.
\newblock Model card. Accessed: Mar 2026.

\bibitem[{Google DeepMind}(2025{\natexlab{b}})]{google2025veo3}
{Google DeepMind}.
\newblock Veo 3: Video generation model with native audio, 2025{\natexlab{b}}.
\newblock URL \url{https://deepmind.google/technologies/veo/}.
\newblock Technical report. Accessed: Mar 2026.

\bibitem[Guo et~al.(2023)Guo, Yang, Rao, Wang, Qiao, Lin, and Dai]{Guo2023AnimateDiffAY}
Yuwei Guo, Ceyuan Yang, Anyi Rao, Yaohui Wang, Y.~Qiao, Dahua Lin, and Bo~Dai.
\newblock Animatediff: Animate your personalized text-to-image diffusion models without specific tuning.
\newblock \emph{ArXiv}, abs/2307.04725, 2023.
\newblock URL \url{https://api.semanticscholar.org/CorpusID:259501509}.

\bibitem[Huang et~al.(2025)Huang, Li, He, Zhou, and Shechtman]{huang2025selfforcingbridgingtraintest}
Xun Huang, Zhengqi Li, Guande He, Mingyuan Zhou, and Eli Shechtman.
\newblock Self forcing: Bridging the train-test gap in autoregressive video diffusion, 2025.
\newblock URL \url{https://arxiv.org/abs/2506.08009}.

\bibitem[Jiang et~al.(2025)Jiang, Han, Mao, Zhang, Pan, and Liu]{Jiang2025VACEAV}
Zeyinzi Jiang, Zhen Han, Chaojie Mao, Jingfeng Zhang, Yulin Pan, and Yu~Liu.
\newblock Vace: All-in-one video creation and editing.
\newblock \emph{ArXiv}, abs/2503.07598, 2025.
\newblock URL \url{https://api.semanticscholar.org/CorpusID:276928131}.

\bibitem[Kahatapitiya et~al.(2025)Kahatapitiya, Ranasinghe, Park, and Ryoo]{kahatapitiya2025language}
Kumara Kahatapitiya, Kanchana Ranasinghe, Jongwoo Park, and Michael~S Ryoo.
\newblock Language repository for long video understanding.
\newblock In \emph{Findings of the Association for Computational Linguistics: ACL 2025}, pages 5627--5646, 2025.

\bibitem[Kalarani et~al.(2024)Kalarani, Bhattacharyya, and Shekhar]{kalarani2024seeing}
Abisek~Rajakumar Kalarani, Pushpak Bhattacharyya, and Sumit Shekhar.
\newblock Seeing the unseen: Visual metaphor captioning for videos.
\newblock \emph{ArXiv, abs/2406.04886}, 1, 2024.

\bibitem[Kim et~al.(2024)Kim, Choi, Lee, and Rhee]{kim2024image}
Wonkyun Kim, Changin Choi, Wonseok Lee, and Wonjong Rhee.
\newblock An image grid can be worth a video: Zero-shot video question answering using a vlm.
\newblock \emph{IEEE Access}, 12:\penalty0 193057--193075, 2024.

\bibitem[Li et~al.(2023)Li, Chong, Stepputtis, Campbell, Hughes, Lewis, and Sycara]{li2023theory}
Huao Li, Yu~Chong, Simon Stepputtis, Joseph~P Campbell, Dana Hughes, Charles Lewis, and Katia Sycara.
\newblock Theory of mind for multi-agent collaboration via large language models.
\newblock In \emph{Proceedings of the 2023 Conference on Empirical Methods in Natural Language Processing}, pages 180--192, 2023.

\bibitem[Li et~al.(2025)Li, Song, Bai, Liang, Yang, Jin, and Mao]{Li2025ICEffectPA}
Yuanhang Li, Yiren Song, Junzhe Bai, Xinran Liang, Hu~Yang, Libiao Jin, and Qi~Mao.
\newblock Ic-effect: Precise and efficient video effects editing via in-context learning.
\newblock \emph{ArXiv}, abs/2512.15635, 2025.
\newblock URL \url{https://api.semanticscholar.org/CorpusID:283920206}.

\bibitem[Lin et~al.(2023{\natexlab{a}})Lin, Zala, Cho, and Bansal]{Lin2023VideoDirectorGPTCM}
Han Lin, Abhaysinh Zala, Jaemin Cho, and Mohit Bansal.
\newblock Videodirectorgpt: Consistent multi-scene video generation via llm-guided planning.
\newblock \emph{ArXiv}, abs/2309.15091, 2023{\natexlab{a}}.
\newblock URL \url{https://api.semanticscholar.org/CorpusID:262825203}.

\bibitem[Lin et~al.(2023{\natexlab{b}})Lin, Ahmed, Li, Lin, Azarnasab, Yang, Wang, Liang, Liu, Lu, et~al.]{lin2023mm}
Kevin Lin, Faisal Ahmed, Linjie Li, Chung-Ching Lin, Ehsan Azarnasab, Zhengyuan Yang, Jianfeng Wang, Lin Liang, Zicheng Liu, Yumao Lu, et~al.
\newblock Mm-vid: Advancing video understanding with gpt-4v (ision).
\newblock \emph{arXiv preprint arXiv:2310.19773}, 2023{\natexlab{b}}.

\bibitem[Lin and Shou(2025)]{lin2025vlog}
Kevin~Qinghong Lin and Mike~Zheng Shou.
\newblock Vlog: Video-language models by generative retrieval of narration vocabulary.
\newblock In \emph{Proceedings of the IEEE/CVF Conference on Computer Vision and Pattern Recognition}, pages 3218--3228, 2025.

\bibitem[Ma et~al.(2025)Ma, He, Wang, Wang, Shen, Qi, Ying, Cai, Li, yeung Shum, Liu, and Chen]{Ma2025FollowYourClickOR}
Yue Ma, Yin-Yin He, Hongfa Wang, Andong Wang, Leqi Shen, Chenyang Qi, Jixuan Ying, Chengfei Cai, Zhifeng Li, Heung yeung Shum, Wei Liu, and Qifeng Chen.
\newblock Follow-your-click: Open-domain regional image animation via motion prompts.
\newblock In \emph{AAAI Conference on Artificial Intelligence}, 2025.
\newblock URL \url{https://api.semanticscholar.org/CorpusID:277751109}.

\bibitem[Mahon and Lapata(2024)]{mahon2024screenwriter}
Louis Mahon and Mirella Lapata.
\newblock Screenwriter: Automatic screenplay generation and movie summarisation.
\newblock \emph{arXiv preprint arXiv:2410.19809}, 2024.

\bibitem[Min et~al.(2024)Min, Buch, Nagrani, Cho, and Schmid]{min2024morevqa}
Juhong Min, Shyamal Buch, Arsha Nagrani, Minsu Cho, and Cordelia Schmid.
\newblock Morevqa: Exploring modular reasoning models for video question answering.
\newblock In \emph{Proceedings of the IEEE/CVF Conference on Computer Vision and Pattern Recognition}, pages 13235--13245, 2024.

\bibitem[Ortega et~al.(2020)Ortega, Kose, Ca{\~n}as, Chao, Unnervik, Nieto, Otaegui, and Salgado]{Ortega2020DMDAL}
Juan~Diego Ortega, Neslihan Kose, Paola~Natalia Ca{\~n}as, Min-An Chao, Alexander Unnervik, Marcos Nieto, Oihana Otaegui, and Luis Salgado.
\newblock Dmd: A large-scale multi-modal driver monitoring dataset for attention and alertness analysis.
\newblock In \emph{ECCV Workshops}, 2020.
\newblock URL \url{https://api.semanticscholar.org/CorpusID:221341025}.

\bibitem[Pan et~al.(2025)Pan, Lu, Wang, Zheng, Wen, Feng, Zhu, and Chen]{pan2025agentcoord}
Bo~Pan, Jiaying Lu, Ke~Wang, Li~Zheng, Zhen Wen, Yingchaojie Feng, Minfeng Zhu, and Wei Chen.
\newblock Agentcoord: Visually exploring coordination strategy for llm-based multi-agent collaboration.
\newblock \emph{Computers \& graphics}, page 104338, 2025.

\bibitem[Po et~al.(2025)Po, Nitzan, Zhang, Chen, Dao, Shechtman, Wetzstein, and Huang]{Po2025LongContextSV}
Ryan Po, Yotam Nitzan, Richard Zhang, Berlin Chen, Tri Dao, Eli Shechtman, Gordon Wetzstein, and Xun Huang.
\newblock Long-context state-space video world models.
\newblock \emph{ArXiv}, abs/2505.20171, 2025.
\newblock URL \url{https://api.semanticscholar.org/CorpusID:278911218}.

\bibitem[Rissanen(1978)]{rissanen1978modeling}
Jorma Rissanen.
\newblock Modeling by shortest data description.
\newblock \emph{Automatica}, 14\penalty0 (5):\penalty0 465--471, 1978.

\bibitem[Runway(2025)]{runway2025gen4}
Runway.
\newblock Introducing runway gen-4.
\newblock \url{https://runwayml.com/research/introducing-runway-gen-4}, 2025.
\newblock Accessed: 2026-03-06.

\bibitem[Song et~al.(2025{\natexlab{a}})Song, Song, Peng, Gao, and Shou]{song2025worldwander}
Quanjian Song, Yiren Song, Kelly Peng, Yuan Gao, and Mike~Zheng Shou.
\newblock Worldwander: Bridging egocentric and exocentric worlds in video generation.
\newblock \emph{arXiv preprint arXiv:2511.22098}, 2025{\natexlab{a}}.

\bibitem[Song et~al.(2024)Song, Huang, Yao, Ye, Ci, Liu, Zhang, and Shou]{song2024processpainter}
Yiren Song, Shijie Huang, Chen Yao, Xiaojun Ye, Hai Ci, Jiaming Liu, Yuxuan Zhang, and Mike~Zheng Shou.
\newblock Processpainter: Learn painting process from sequence data.
\newblock \emph{arXiv preprint arXiv:2406.06062}, 2024.

\bibitem[Song et~al.(2025{\natexlab{b}})Song, Liu, Mao, and Shou]{song2025mitty}
Yiren Song, Cheng Liu, Weijia Mao, and Mike~Zheng Shou.
\newblock Mitty: Diffusion-based human-to-robot video generation.
\newblock \emph{arXiv preprint arXiv:2512.17253}, 2025{\natexlab{b}}.

\bibitem[Sur{\'\i}s et~al.(2023)Sur{\'\i}s, Menon, and Vondrick]{suris2023vipergpt}
D{\'\i}dac Sur{\'\i}s, Sachit Menon, and Carl Vondrick.
\newblock Vipergpt: Visual inference via python execution for reasoning.
\newblock In \emph{Proceedings of the IEEE/CVF international conference on computer vision}, pages 11888--11898, 2023.

\bibitem[Tang et~al.(2025)Tang, Bi, Xu, Song, Liang, Wang, Zhang, An, Lin, Zhu, et~al.]{tang2025video}
Yunlong Tang, Jing Bi, Siting Xu, Luchuan Song, Susan Liang, Teng Wang, Daoan Zhang, Jie An, Jingyang Lin, Rongyi Zhu, et~al.
\newblock Video understanding with large language models: A survey.
\newblock \emph{IEEE Transactions on Circuits and Systems for Video Technology}, 2025.

\bibitem[Team et~al.(2025)Team, Chen, Ci, Du, Feng, Gai, Guo, Han, He, He, et~al.]{team2025kling}
Kling Team, Jialu Chen, Yuanzheng Ci, Xiangyu Du, Zipeng Feng, Kun Gai, Sainan Guo, Feng Han, Jingbin He, Kang He, et~al.
\newblock Kling-omni technical report.
\newblock \emph{arXiv preprint arXiv:2512.16776}, 2025.

\bibitem[Tran et~al.(2025)Tran, Dao, Nguyen, Pham, O'Sullivan, and Nguyen]{tran2025multi}
Khanh-Tung Tran, Dung Dao, Minh-Duong Nguyen, Quoc-Viet Pham, Barry O'Sullivan, and Hoang~D Nguyen.
\newblock Multi-agent collaboration mechanisms: A survey of llms.
\newblock \emph{arXiv preprint arXiv:2501.06322}, 2025.

\bibitem[Wang et~al.(2025{\natexlab{a}})Wang, Ai, Wen, Mao, Xie, Chen, Yu, Zhao, Yang, Zeng, Wang, Zhang, Zhou, Wang, Chen, Zhu, Zhao, Yan, Huang, Meng, Zhang, Li, Wu, Chu, Feng, Zhang, Sun, Fang, Wang, Gui, Weng, Shen, Lin, Wang, Wang, Zhou, Wang, Shen, Yu, Shi, Huang, Xu, Kou, Lv, Li, Liu, Wang, Zhang, Huang, Li, Wu, Liu, Pan, Zheng, Hong, Shi, Feng, Jiang, Han, Wu, and Liu]{Wang2025WanOA}
Ang Wang, Baole Ai, Bin Wen, Chaojie Mao, Chen-Wei Xie, Di~Chen, Feiwu Yu, Haiming Zhao, Jianxiao Yang, Jianyuan Zeng, Jiayu Wang, Jingfeng Zhang, Jingren Zhou, Jinkai Wang, Jixuan Chen, Kai Zhu, Kang Zhao, Keyu Yan, Lianghua Huang, Xiaofeng Meng, Ningying Zhang, Pandeng Li, Ping Wu, Ruihang Chu, Rui Feng, Shiwei Zhang, Siyang Sun, Tao Fang, Tianxing Wang, Tianyi Gui, Tingyu Weng, Tong Shen, Wei Lin, Wei Wang, Wei Wang, Wen-Chao Zhou, Wente Wang, Wen Shen, Wenyuan Yu, Xianzhong Shi, Xiaomin Huang, Xin Xu, Yan Kou, Yan-Mei Lv, Yifei Li, Yijing Liu, Yiming Wang, Yingya Zhang, Yitong Huang, Yong Li, You Wu, Yu~Liu, Yulin Pan, Yun Zheng, Yuntao Hong, Yupeng Shi, Yutong Feng, Zeyinzi Jiang, Zhen Han, Zhigang Wu, and Ziyu Liu.
\newblock Wan: Open and advanced large-scale video generative models.
\newblock \emph{ArXiv}, abs/2503.20314, 2025{\natexlab{a}}.
\newblock URL \url{https://api.semanticscholar.org/CorpusID:277321639}.

\bibitem[Wang et~al.(2025{\natexlab{b}})Wang, Shi, Li, Bian, Liu, Lu, Wang, Wan, Gai, and Jia]{wang2025multishotmaster}
Qinghe Wang, Xiaoyu Shi, Baolu Li, Weikang Bian, Quande Liu, Huchuan Lu, Xintao Wang, Pengfei Wan, Kun Gai, and Xu~Jia.
\newblock Multishotmaster: A controllable multi-shot video generation framework.
\newblock \emph{arXiv preprint arXiv:2512.03041}, 2025{\natexlab{b}}.

\bibitem[Wang et~al.(2024)Wang, Mao, Wu, Ge, Wei, and Ji]{wang2024unleashing}
Zhenhailong Wang, Shaoguang Mao, Wenshan Wu, Tao Ge, Furu Wei, and Heng Ji.
\newblock Unleashing the emergent cognitive synergy in large language models: A task-solving agent through multi-persona self-collaboration.
\newblock In \emph{Proceedings of the 2024 Conference of the North American Chapter of the Association for Computational Linguistics: Human Language Technologies (Volume 1: Long Papers)}, pages 257--279, 2024.

\bibitem[Wiedemer et~al.(2025)Wiedemer, Li, Vicol, Gu, Matarese, Swersky, Kim, Jaini, and Geirhos]{Wiedemer2025VideoMA}
Thaddaus Wiedemer, Yuxuan Li, Paul Vicol, Shixiang~Shane Gu, Nick Matarese, Kevin Swersky, Been Kim, Priyank Jaini, and Robert Geirhos.
\newblock Video models are zero-shot learners and reasoners.
\newblock \emph{ArXiv}, abs/2509.20328, 2025.
\newblock URL \url{https://api.semanticscholar.org/CorpusID:281505752}.

\bibitem[Wu et~al.(2025{\natexlab{a}})Wu, Zou, Li, Huang, Yang, Tan, Peng, Wu, Xiong, Jiang, Linus, Patrol, Zhang, Chen, Zhao, Tian, Liu, Kong, Wang, He, Li, Deng, Zhe, Li, Long, Peng, Wu, Liu, Wang, Dai, Peng, Li, Gong, Xiao, Tian, Lin, Liu, Zhang, Lian, Pan, Wang, Niu, Chen, Chen, Zheng, Yang, Hu, Yang, Xiao, Wu, Xu, Yuan, Sang, Huang, Gong, Huang, Guo, Yuan, Chen, Hu, Sun, Wu, Ren, Yuan, Mi, Zhang, Sun, Lu, Li, Huang, Tang, Li, Deng, Zhou, Hu, Liu, Yang, Yang, Lu, Zhou, and Zhong]{wu2025hunyuanvideo15technicalreport}
Bing Wu, Chang Zou, Changlin Li, Duojun Huang, Fang Yang, Hao Tan, Jack Peng, Jianbing Wu, Jiangfeng Xiong, Jie Jiang, Linus, Patrol, Peizhen Zhang, Peng Chen, Penghao Zhao, Qi~Tian, Songtao Liu, Weijie Kong, Weiyan Wang, Xiao He, Xin Li, Xinchi Deng, Xuefei Zhe, Yang Li, Yanxin Long, Yuanbo Peng, Yue Wu, Yuhong Liu, Zhenyu Wang, Zuozhuo Dai, Bo~Peng, Coopers Li, Gu~Gong, Guojian Xiao, Jiahe Tian, Jiaxin Lin, Jie Liu, Jihong Zhang, Jiesong Lian, Kaihang Pan, Lei Wang, Lin Niu, Mingtao Chen, Mingyang Chen, Mingzhe Zheng, Miles Yang, Qiangqiang Hu, Qi~Yang, Qiuyong Xiao, Runzhou Wu, Ryan Xu, Rui Yuan, Shanshan Sang, Shisheng Huang, Siruis Gong, Shuo Huang, Weiting Guo, Xiang Yuan, Xiaojia Chen, Xiawei Hu, Wenzhi Sun, Xiele Wu, Xianshun Ren, Xiaoyan Yuan, Xiaoyue Mi, Yepeng Zhang, Yifu Sun, Yiting Lu, Yitong Li, You Huang, Yu~Tang, Yixuan Li, Yuhang Deng, Yuan Zhou, Zhichao Hu, Zhiguang Liu, Zhihe Yang, Zilin Yang, Zhenzhi Lu, Zixiang Zhou, and Zhao Zhong.
\newblock Hunyuanvideo 1.5 technical report, 2025{\natexlab{a}}.
\newblock URL \url{https://arxiv.org/abs/2511.18870}.

\bibitem[Wu et~al.(2022)Wu, Ge, Wang, Lei, Gu, Hsu, Shan, Qie, and Shou]{Wu2022TuneAVideoOT}
Jay~Zhangjie Wu, Yixiao Ge, Xintao Wang, Weixian Lei, Yuchao Gu, Wynne Hsu, Ying Shan, Xiaohu Qie, and Mike~Zheng Shou.
\newblock Tune-a-video: One-shot tuning of image diffusion models for text-to-video generation.
\newblock \emph{2023 IEEE/CVF International Conference on Computer Vision (ICCV)}, pages 7589--7599, 2022.
\newblock URL \url{https://api.semanticscholar.org/CorpusID:254974187}.

\bibitem[Wu et~al.(2025{\natexlab{b}})Wu, Zhu, and Shou]{wu2025automated}
Weijia Wu, Zeyu Zhu, and Mike~Zheng Shou.
\newblock Automated movie generation via multi-agent cot planning.
\newblock \emph{arXiv preprint arXiv:2503.07314}, 2025{\natexlab{b}}.

\bibitem[Xu et~al.(2026)Xu, Ma, Wang, Peng, Liang, and Li]{xu2026mochaendtoendvideocharacterreplacement}
Zhengbo Xu, Jie Ma, Ziheng Wang, Zhan Peng, Jun Liang, and Jing Li.
\newblock Mocha:end-to-end video character replacement without structural guidance, 2026.
\newblock URL \url{https://arxiv.org/abs/2601.08587}.

\bibitem[Xu et~al.(2025)Xu, Wang, Wang, Li, Shi, Yang, Wang, Hu, Yu, and Zhang]{xu2025filmagentmultiagentframeworkendtoend}
Zhenran Xu, Longyue Wang, Jifang Wang, Zhouyi Li, Senbao Shi, Xue Yang, Yiyu Wang, Baotian Hu, Jun Yu, and Min Zhang.
\newblock Filmagent: A multi-agent framework for end-to-end film automation in virtual 3d spaces, 2025.
\newblock URL \url{https://arxiv.org/abs/2501.12909}.

\bibitem[Yan et~al.(2026)Yan, Ye, Yang, Teng, Dong, Wen, Gu, Liu, and Tang]{yan2026scailstudiogradecharacteranimation}
Wenhao Yan, Sheng Ye, Zhuoyi Yang, Jiayan Teng, ZhenHui Dong, Kairui Wen, Xiaotao Gu, Yong-Jin Liu, and Jie Tang.
\newblock Scail: Towards studio-grade character animation via in-context learning of 3d-consistent pose representations, 2026.
\newblock URL \url{https://arxiv.org/abs/2512.05905}.

\bibitem[Yang et~al.(2025)Yang, Ci, Song, and Shou]{yang2025x}
Pei Yang, Hai Ci, Yiren Song, and Mike~Zheng Shou.
\newblock X-humanoid: Robotize human videos to generate humanoid videos at scale.
\newblock \emph{arXiv preprint arXiv:2512.04537}, 2025.

\bibitem[Zeng et~al.(2025)Zeng, Cai, Chen, Lv, and Wang]{zeng2025coagentcollaborativeplanningconsistency}
Qinglin Zeng, Kaitong Cai, Ruiqi Chen, Qinhan Lv, and Keze Wang.
\newblock Coagent: Collaborative planning and consistency agent for coherent video generation, 2025.
\newblock URL \url{https://arxiv.org/abs/2512.22536}.

\bibitem[Zhang et~al.(2024)Zhang, Jiao, Chen, Zhao, Tan, Li, Ma, and Chen]{zhang2024eventhallusion}
Jiacheng Zhang, Yang Jiao, Shaoxiang Chen, Na~Zhao, Zhiyu Tan, Hao Li, Xingjun Ma, and Jingjing Chen.
\newblock Eventhallusion: Diagnosing event hallucinations in video llms.
\newblock \emph{arXiv preprint arXiv:2409.16597}, 2024.

\bibitem[Zhang et~al.(2025{\natexlab{a}})Zhang, Jiang, Wang, Fang, Zhi, Yan, Kang, Lu, and Pan]{zhang2025storymem}
Kaiwen Zhang, Liming Jiang, Angtian Wang, Jacob~Zhiyuan Fang, Tiancheng Zhi, Qing Yan, Hao Kang, Xin Lu, and Xingang Pan.
\newblock Storymem: Multi-shot long video storytelling with memory.
\newblock \emph{arXiv preprint arXiv:2512.19539}, 2025{\natexlab{a}}.

\bibitem[Zhang et~al.(2025{\natexlab{b}})Zhang, Xu, Yang, Yin, Liu, Xu, Wang, Wu, Hong, Zhang, Liang, and Jiang]{zhang2025animeadaptivemultiagentplanning}
Lisai Zhang, Baohan Xu, Siqian Yang, Mingyu Yin, Jing Liu, Chao Xu, Siqi Wang, Yidi Wu, Yuxin Hong, Zihao Zhang, Yanzhang Liang, and Yudong Jiang.
\newblock Anime: Adaptive multi-agent planning for long animation generation, 2025{\natexlab{b}}.
\newblock URL \url{https://arxiv.org/abs/2508.18781}.

\bibitem[Zhang et~al.(2025{\natexlab{c}})Zhang, Cai, Li, Wetzstein, and Agrawala]{Zhang2025FrameCP}
Lvmin Zhang, Shengqu Cai, Muyang Li, Gordon Wetzstein, and Maneesh Agrawala.
\newblock Frame context packing and drift prevention in next-frame-prediction video diffusion models.
\newblock 2025{\natexlab{c}}.
\newblock URL \url{https://api.semanticscholar.org/CorpusID:277857265}.

\bibitem[Zhao et~al.(2023)Zhao, Misra, Kr{\"a}henb{\"u}hl, and Girdhar]{zhao2023learning}
Yue Zhao, Ishan Misra, Philipp Kr{\"a}henb{\"u}hl, and Rohit Girdhar.
\newblock Learning video representations from large language models.
\newblock In \emph{Proceedings of the IEEE/CVF conference on computer vision and pattern recognition}, pages 6586--6597, 2023.

\end{thebibliography}

%%%%%%%%%%%%%%%%%%%%%%%%%%%%%%%%%%%%%%%%%%%%%%%%%%%%%%%%%%%%

\appendix

% =========================
% Supplementary Material
\clearpage
\setcounter{page}{1}
\clearpage
\setcounter{page}{1}

% 重置章节计数器，让附录从A开始编号
\setcounter{section}{0}
\renewcommand{\thesection}{\Alph{section}}

% \renewcommand{\headrulewidth}{0pt}

% 补充材料标题 - 居中显示
\begin{center}
\Large\textbf{Appendix}
\end{center}

The appendix is organized as follows:

\begin{enumerate}
    \item \textbf{Section A: User Study} - We provide details of our user study protocol for evaluating series-level cinematic remaking, including the evaluation criteria, compared methods, questionnaire design, and preference results.

    \item \textbf{Section B: Multi-Agent System Implementation Details} - We provide comprehensive implementation details of our three collaborative agents: (1) Video Understanding Agent for structured screenplay generation and character tracking, (2) Video Generation Agent for anchor-driven keyframe and video synthesis, and (3) Verification Agent for closed-loop quality auditing. We also detail the contextual memory allocation mechanism and reference generation process that support the Dual-Bridge Consistency framework.

    \item \textbf{Section C: Output JSON Format} - We present the structured output format from our video understanding module, demonstrating comprehensive JSON examples including top-level organization, character roster structure, and shot-level analysis with technical cinematic parameters and narrative descriptions.

    \item \textbf{Section D: Grid Joint Synthesis Visualization} - We demonstrate our grid joint synthesis strategy for improving intra-scene consistency, including $2\times2$ and $3\times3$ grid generation examples.

    \item \textbf{Section E: More Keyframe Results} - We present 7 additional comprehensive visual comparisons across videos ranging from 14 to over 40 shots, providing keyframe-by-keyframe analysis that demonstrates the visual fidelity and consistency of our generation results.

    % \item \textbf{Section F: Baseline Comparisons with Video Generation Models} - We report comprehensive evaluations against state-of-the-art video generation baselines including Kling, VEO3, and Batch Video Generation, with both quantitative metrics and qualitative comparisons.
\end{enumerate}

% \section{User Study}
% \label{sec:agent-details}

% We conduct a user study to complement automated evaluations and assess perceptual quality in series-level cinematic remaking. 
% We evaluate four criteria: \textit{Character Consistency}, \textit{Scene Consistency}, \textit{Plot Consistency}, and \textit{Overall Preference}. We collect 20 anonymous questionnaires from participants with diverse backgrounds. Each questionnaire presents 10 groups of remade samples. 
% Each group contains results generated by \textbf{Mocha}, \textbf{Kling O1}, \textbf{Runway Gen4}, and our method. 
% For each evaluation criterion, participants are asked to select the method they prefer most among the four candidates.

% As shown in Fig.~\ref{user_study}, our method achieves the highest user preference across all four criteria.

% \begin{figure}[h]
%   \centering
%   \includegraphics[width=\linewidth]{image/user_study_chart_flat.pdf}
%   \caption{User study preference distribution across four criteria: identity consistency, scene consistency, plot consistency, and overall preference.}
%   \label{user_study}
% \end{figure}

\section{User Study}
\label{sec:user-study}

We conduct a user study to complement automated evaluations and assess perceptual quality in series-level cinematic remaking. 
We evaluate four criteria: \textit{Character Consistency}, \textit{Scene Consistency}, \textit{Plot Consistency}, and \textit{Overall Preference}. 
We collect 20 anonymous questionnaires from participants with experience in AI-based film, video generation, or video production, including students and practitioners. 
Each questionnaire presents 10 groups of remade samples. 
For each group, participants are shown the outputs of \textbf{Mocha}, \textbf{Kling O1}, \textbf{Runway Gen4}, and our method in a randomized and anonymized interface, without revealing which method produced each result. 
Participants can view all candidate videos side by side and are asked to select the best result under each evaluation criterion. 
This blind comparison protocol reduces method-name bias and focuses the evaluation on perceived identity stability, scene coherence, plot fidelity, and overall quality.

\begin{figure}[h]
  \centering
  \includegraphics[width=\linewidth]{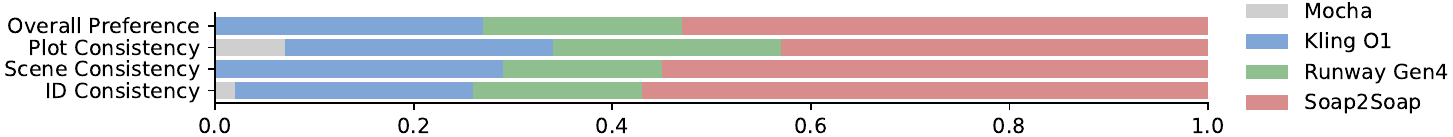}
  \caption{User study preference distribution across four criteria: identity consistency, scene consistency, plot consistency, and overall preference.}
  \label{user_study}
\end{figure}

As shown in Fig.~\ref{user_study}, our method achieves the highest user preference across all four criteria.

%==============================================================================
\section{Multi-Agent System Implementation Details}
\label{sec:agent-details}

This section provides detailed implementation specifics for the three collaborative agents in the Soap2Soap framework. The system follows the Dual-Bridge Consistency design principle introduced in the main paper, achieving both semantic fidelity and visual coherence through the coordinated operation of the Language Bridge and Visual Bridge.

%------------------------------------------------------------------------------
\subsection{Video Understanding Agent}
\label{sec:video-understanding}

The Video Understanding Agent serves as the entry point of the entire pipeline, transforming unstructured video content into the structured JSON screenplay $S_{\text{json}}$ that functions as the Language Bridge throughout the processing flow. This agent employs a multi-stage prompt engineering strategy for global character identification and tracking. By fusing information from three modalities—on-screen label recognition, dialogue content analysis, and visual feature matching—the system constructs multi-dimensional character profiles encompassing primary names, aliases, titles, role classifications, and familial relationships. Physical attributes including gender, age range, body type, skin tone, hairstyle, and facial characteristics are also encoded to establish cross-scene identity consistency constraints.

For scene understanding, the agent adopts an adaptive scene boundary detection algorithm to automatically identify scene transitions in the video. For each detected scene, the system generates environment-only descriptions that strictly exclude character information, focusing solely on spatial layout, lighting configuration, color atmosphere, and temporal attributes. This design strategy ensures the purity of environment reference generation, avoiding interference from character features in background synthesis, thereby effectively supporting scene-level stability through the Visual Bridge.

A critical component of the Video Understanding Agent is the fine-grained wardrobe DNA extraction system. The system introduces a 7-dimension clothing specification framework that comprehensively characterizes costume details from multiple perspectives: the color system employs dual Pantone TCX and HEX encoding for precise color matching; the fabric system encompasses eight attributes including material, weave pattern, weight, opacity, finish, stretch, texture, and drape; the cut \& fit system records tailoring details for both tops and bottoms; the details system describes decorative elements such as buttons, zippers, pockets, and stitching; the pattern system captures pattern type, size, arrangement, direction, and density; the accessories system documents footwear, jewelry, and bags; and the styling system describes layering, tucking, sleeve state, and overall style. This fine-grained feature representation enables precise control over costume visual presentation during generation, effectively preventing detail loss or color drift.

For each shot $s_{i,j}$, the agent extracts two categories of critical information to support subsequent generation. The narrative description component includes T2I prompts for static keyframe generation and I2V prompts for dynamic video generation, which not only describe visual content but also explicitly specify action flow, camera movement, and temporal progression. The technical parameter component covers multiple dimensions of cinematography including lighting setup, color grading, composition, mood atmosphere, shot size, camera angle, camera height, horizontal angle, focal length, depth of field, technical device metadata, and camera movement. These parameters ensure that generated content accurately reproduces the visual style and cinematic language of the original video.

%------------------------------------------------------------------------------
\subsection{Contextual Memory Allocation}
\label{sec:memory-allocation}

The contextual memory allocation mechanism represents a core functionality within the Video Understanding Agent, specifically addressing the context management challenges posed by multiple characters and scenes in long videos. In long video processing, directly loading the entire global context for every shot is both redundant and unstable, frequently leading to role confusion and visual quality degradation. Therefore, the system adopts a dynamic memory package construction strategy. For each shot $s_{i,j}$, the system dynamically constructs a memory package $\mathcal{M}_{i,j}$ based on the narrative context in $S_{\text{json}}$. The construction follows the principle of minimal sufficiency, containing only the minimal necessary information required to generate the current shot, thereby avoiding the loading of irrelevant history.

Each memory package $\mathcal{M}_{i,j}$ comprises two complementary components. The semantic instruction component provides narrative information including shot description, T2I prompts, I2V prompts, and character action sequences, extracted directly from $S_{\text{json}}$ and organized in a format comprehensible to generation models. The visual reference anchor component supplies three types of visual constraints: character identity references (including character ID, facial features, and appearance cues), clothing references (clothing reference images generated based on the 7-dimension DNA description), and scene environment references (scene-level background layout, lighting, and style images). This dual-component design ensures that the generation process is guided by both semantic instructions and visual constraints.

Memory allocation is organized at two granularities: scene level and shot level. At the scene level, the system allocates environment reference images for each major scene, which are shared across all shots within that scene, ensuring consistency in background layout, lighting conditions, and overall style. At the shot level, the system dynamically allocates character identity references and clothing references based on specific shot content, adapting to character appearances and disappearances across different shots. This dual-level allocation strategy effectively balances scene consistency with shot flexibility, avoiding the role confusion and background mutation issues that global reference sets encounter in multi-character, multi-scene videos.

\textbf{Memory Package Structure:}
A simplified example of the memory package structure is illustrated below:
\begin{lstlisting}[style=jsonstyle]
{
  "shot_id": "9",
  "major_scene": "major_scene_01",
  "characters": ["@character_01", "@character_02"],
  "environment_ref": "reference_images/major_scene_01_environment.png",
  "clothing_refs": ["reference_images/major_scene_01_character_01_clothing.png"],
  "character_refs": ["reference_images/character_01_portrait.png"],
  "visual_dna": { /* lighting, color, mood, etc */ },
  "narrative": { /* action, camera_movement, language_prompt */ },
  "character_mappings": { /* char_id -> target_name, clothing */ },
  "style_prompt": "STYLE: REALISTIC CINEMATIC...",
  "generation_feedback": []
}
\end{lstlisting}

%------------------------------------------------------------------------------
\subsection{Reference Generation}
\label{sec:reference-gen}

Before keyframe generation, the system first generates two categories of reference images that serve as concrete carriers of the Visual Bridge. Environment reference generation strictly follows the environment-only principle, containing only spatial structure, lighting atmosphere, and background elements of the scene while maintaining the original video's aspect ratio (16:9 or 9:16). A style-aware generation strategy is employed to ensure consistency with the target artistic style. Clothing reference generation is more complex, requiring the integration of target character facial features with scene-specific costume configurations to generate complete full-body reference images. For 2D animation styles (such as Family Guy or LEGO), the system adopts a special copy mode that directly reuses character designs from the original style to ensure stylistic unity. These reference images continuously provide visual constraints during subsequent generation, effectively suppressing character identity drift and scene background mutation.

%------------------------------------------------------------------------------
\subsection{Anchor-Driven Visual Generation}
\label{sec:generation}

The Video Generation Agent performs visual synthesis based on semantic instructions and visual anchors provided by memory packages. To enhance intra-scene consistency, particularly when handling large viewpoint variation scenarios such as reverse shot sequences, the system adopts a grid joint synthesis strategy. Specifically, the system organizes 4 or 9 shots sharing the same scene and character configuration into a $2\times2$ or $3\times3$ grid structure, jointly generating multiple keyframes in a single generation call. The core advantage of this design is that all sub-images share self-attention mechanisms within the same generation context, thereby forming stronger identity consistency and scene coherence within the grid. The generated grid images are subsequently decomposed into individual keyframes for downstream video synthesis.

Given the generated keyframes, the system employs image-to-video synthesis technology to produce 4 to 8 second video clips. The system calls the Veo 3 model, inputting scene references, character references, and keyframes from the memory package as multiple reference images, while simultaneously providing I2V prompts extracted from $S_{\text{json}}$ as dynamic guidance. This prompt explicitly specifies camera movement, character actions, and shot dynamics, ensuring that generated video clips faithfully reproduce the original video's motion choreography. All shot clips are ultimately concatenated in temporal order to form the complete remade video.

During the generation process, the system employs an enhanced prompt construction strategy that organically integrates visual DNA, narrative context, reference image paths, and style prompts from the memory package. The system strictly enforces technical specification constraints including precise aspect ratio preservation, active text element elimination (subtitles, watermarks, overlays), and generation quality control. Reference image integration follows a priority mechanism: environment references have the highest priority, followed by clothing references, and finally character portrait references. The number of character portrait references is limited to a maximum of 5 per generation to avoid quality degradation from information overload.

%------------------------------------------------------------------------------
\subsection{Verification Agent}
\label{sec:verification}

The Verification Agent serves as the closed-loop feedback mechanism of the system, conducting multi-dimensional quality audits on generated content and triggering selective regeneration upon detecting consistency defects. For each shot $s_{i,j}$, the Verification Agent evaluates generated keyframes and video clips across four dimensions: Generation Quality verifies artifact-free rendering and correct character counts consistent with $S_{\text{json}}$; Identity \& Appearance matches facial IDs and clothing features in generated content against visual anchors; Environmental \& Style Stability checks whether backgrounds and styles remain consistent with scene reference anchors; and Plot Consistency verifies that character actions and interactions faithfully reflect the narrative descriptions in $S_{\text{json}}$.

When inconsistencies are detected (such as identity drift, attribute mismatch, or scene inconsistency), the Verification Agent generates structured textual feedback $\Delta$. This feedback is routed to the Video Understanding and Video Generation Agents to optimize screenplay descriptions and spatial control signals respectively. The context update process adjusts the memory package and I2V prompts based on feedback, generating updated conditioning $(\mathcal{M}_{i,j}', P_{\text{I2V}}')$. The Video Generation Agent then performs regeneration for affected shots based on the updated conditions. The system supports configurable maximum retry counts to balance generation quality and computational cost. This iterative correction loop continues until all evaluation criteria are met, thereby enforcing long-term consistency without full sequence rollback. This mechanism effectively suppresses error accumulation during long video generation, ensuring system robustness at scales of hundreds of shots.

The Verification Agent employs a dual-mode strategy for face matching. The primary approach utilizes Gemini Vision API for semantic-level facial understanding and matching, suitable for most scenarios. When the primary approach is unavailable or matching scores fall below thresholds, the system automatically switches to a fallback approach using the face\_recognition library for feature-point-based matching. Both approaches output similarity scores that are compared against preset acceptance thresholds to determine whether generated content passes verification. For clothing consistency verification, the agent performs visual comparison between clothing regions in generated images and clothing reference images, with emphasis on color matching (quantitatively assessed based on the Pantone color system), detail consistency (buttons, pockets, stitching), and overall stylistic unity.

%==============================================================================
\section{Output Json Format}
\label{sec:schema}

This section presents the structured output format generated by the Video Understanding Agent. After processing an input video, the agent produces a comprehensive JSON screenplay $S_{\text{json}}$ that serves as the Language Bridge throughout the system. This JSON output organizes extracted information into four main components: (1) \textbf{video metadata} containing technical specifications such as resolution and aspect ratio, (2) \textbf{character roster} providing detailed profiles for all identified characters including names, physical attributes, and clothing variations, (3) \textbf{major scenes} documenting environment descriptions and scene-level attributes, and (4) \textbf{shot-level analysis} containing fine-grained cinematic parameters and narrative descriptions for each individual shot. This hierarchical structure enables efficient context retrieval and supports the contextual memory allocation mechanism described in Section A.

%------------------------------------------------------------------------------
\subsection{Video Metadata}
\label{sec:output-structure}

\begin{lstlisting}[style=jsonstyle]
{
  "video_file": "video.mp4",
  "video_metadata": {
    "aspect_ratio": {
      "width": 1920,
      "height": 1080,
      "aspect_ratio": "16:9",
      "ratio_decimal": 1.777778
    }
  },
  "total_scenes": 20
}
\end{lstlisting}

%------------------------------------------------------------------------------
\subsection{Character Roster Structure}
\label{sec:character-roster}

\begin{lstlisting}[style=jsonstyle]
{
  "characters": [
    {
      "character_id": "@character_01",
      "names": {
        "primary_name": "Emma Smith",
        "primary_source": "on_screen_label",
        "aliases": ["Emma", "Em"],
        "titles": ["Ms. Smith"],
        "roles": ["Protagonist"]
      },
      "physical_attributes": "Female, young adult (20s)",
      "hair": "Red, curly, long",
      "face": "Fair complexion, red lips",
      "clothing_variations": [
        {"scene": "Scene 1", "description": "White lace dress"}
      ],
      "first_appearance": "0.0s",
      "role_classification": "main"
    }
  ]
}
\end{lstlisting}

%------------------------------------------------------------------------------
\subsection{Major Scene Structure}
\label{sec:major-scene}
\begin{lstlisting}[style=jsonstyle]
{
    "major_scenes": [
    {
        "scene_id": "major_scene_01",
        "start_time": 0.0,
        "end_time": 25.0,
        "duration": 25.0,
        "location_type": "Military Hospital Ward",
        "setting_description": "Daytime hospital ward with rows of beds and mosquito nets",
        "lighting_style": "Bright natural daylight through blinds, high key",
        "color_palette": "Institutional green, white, pale yellow",
        "environment_description": "A spacious, utilitarian hospital ward characterized by rows of white metal-framed beds lining both sides of a central aisle. Each bed is draped with sheer white mosquito netting hanging from tall metal frames attached to the headboards and footboards. The walls are painted a pale, institutional green, and the ceiling is high and white. On the right side, a wall of large windows featuring horizontal blinds allows bright, diffuse natural daylight to flood the room, creating a pattern of light and shadow on the light-colored tiled floor. Small metal bedside cabinets and medical stands are positioned next to each bed. The atmosphere is airy and clinical."
    }
}
\end{lstlisting}

\subsection{Shot-Level Analysis Structure}
\label{sec:shot-structure}
Technical Cinematic Parameters and Narrative Descriptions
\begin{lstlisting}[style=jsonstyle]
{
  "shot_id": "9",
  "scene_id": "major_scene_01",
  "lighting_setup": "Three-point lighting with soft key light from camera left",
  "color_grading": "Warm tones, high contrast, cinematic LUT",
  "composition": "Rule of thirds, subject positioned left",
  "mood_atmosphere": "Tense, dramatic atmosphere",
  "shot_size": "Medium close-up",
  "camera_angle": "Eye level",
  "camera_height": "1.5m from ground",
  "horizontal_angle": "15 degrees to subject's left",
  "focal_length": "50mm lens",
  "depth_of_field": "Shallow DOF, background blurred",
  "tech_device": "ARRI Alexa, Cooke S4/i lenses",
  "camera_movement": "Slow push-in, tracking forward",
  "subject_movement": {
    "action": "Character turns head to look at door",
    "dialogue": {
      "timestamp": "00:02:15.500",
      "text": "What was that sound?"
    }
  },
  "I2V_Prompt": "Dynamic sequence: character slowly turns head toward door, expression shifts from neutral to concerned, subtle hand movement, camera pushes forward slightly, 4-second duration",
  "Language_to_One_Shot_Prompt": "Medium close-up of character with concerned expression, looking toward door on frame right, warm lighting from left, shallow depth of field"
}
\end{lstlisting}

%==============================================================================
\section{Grid Joint Synthesis Visualization}
\label{sec:grid-samples}

This section provides visual demonstrations of the grid joint synthesis strategy employed in our anchor-driven keyframe generation process. As described in the main paper, Soap2Soap performs visual rendering in two stages where both keyframe generation and shot-level video synthesis strictly condition on the allocated contextual memory $\mathcal{M}_{i,j}$ to suppress identity drift and scene mutation under frequent shot transitions.

In the keyframe generation stage, we face a significant challenge: maintaining intra-scene consistency when dealing with large viewpoint changes, particularly in scenarios such as reverse shot sequences where the camera angle dramatically shifts between over-the-shoulder shots and reverse angles. Traditional sequential generation approaches, where each keyframe is generated independently, often lead to accumulated inconsistencies in character identity, clothing details, and environmental attributes across shots within the same scene.

To address this challenge, we adopt a grid joint synthesis strategy that fundamentally rethinks how multiple keyframes are generated. Instead of producing each keyframe in isolation, we group 4 or 9 frames that share the same scene and character configuration, and generate them as a single unified canvas in one generation pass. This approach effectively produces a high-resolution keyframe grid—either $2\times2$ or $3\times3$—simultaneously, while enabling all sub-images within the grid to share attention mechanisms within the same generation context. The shared attention is critical because it allows the generation model to establish cross-frame consistency constraints during the denoising process itself, rather than attempting to enforce consistency post-hoc through external verification and regeneration. This intrinsic consistency enforcement leads to significantly stronger identity preservation and scene coherence across all frames within the grid. Additionally, generating multiple keyframes in a single API call improves computational efficiency compared to sequential generation. Once the complete grid is generated, it is decomposed into individual keyframes $I_K$ that serve as the foundation for downstream video synthesis.

The choice between $2\times2$ and $3\times3$ grid configurations depends on the specific scene characteristics and the number of shots requiring consistent generation. The $2\times2$ configuration is particularly effective for shorter scene segments with 4 consecutive shots, providing a balance between consistency enforcement and generation flexibility. The $3\times3$ configuration, on the other hand, is designed for more complex scenarios involving up to 9 shots, such as dialogue-heavy scenes with multiple camera angles and frequent viewpoint reversals. Both configurations leverage the same underlying principle: by generating multiple keyframes within a shared latent space, the model can maintain consistent character appearances, clothing details, and environmental attributes across all frames, even when the camera perspective changes dramatically from shot to shot.

\begin{figure}[h]
\centering
\includegraphics[width=0.95\textwidth]{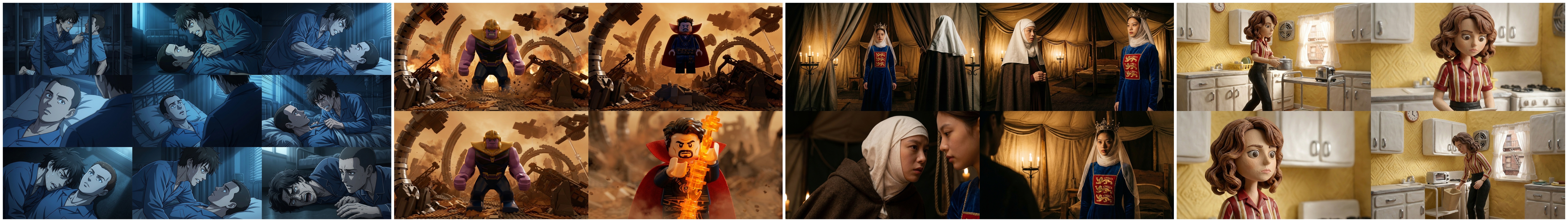}
\caption{\textbf{Grid Joint Synthesis Examples.} }
\label{fig:grid}
\end{figure}

%------------------------------------------------------------------------------
Figure~\ref{fig:grid} presents four grid joint synthesis example demonstrating how our system generates 9 or consecutive keyframes in a single pass. All frames share the same major scene and character configuration, yet exhibit small viewpoint and pose variations that are typical in cinematic sequences.

%------------------------------------------------------------------------------

%------------------------------------------------------------------------------

%==============================================================================
\section{More Keyframe Results}
\label{sec:additional-results}

\begin{figure}[h]
\centering
% TODO: Replace with actual keyframe comparison image 1
\includegraphics[width=0.95\textwidth]{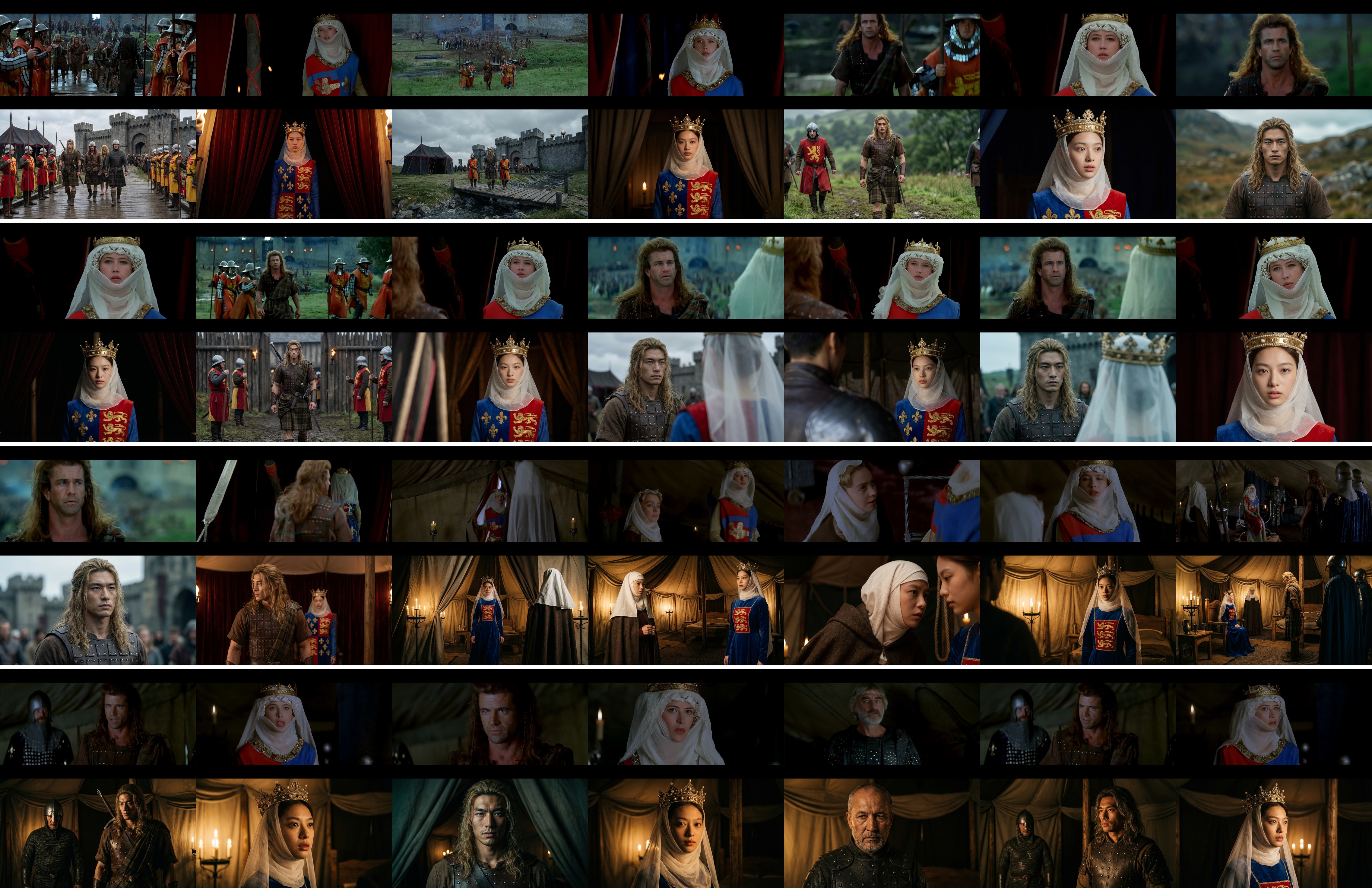}
\caption{\textbf{Braveheart (Style :} Realistic)}
\label{fig:keyframe_result_1}
\end{figure}

\begin{figure}[h]
\centering
% TODO: Replace with actual keyframe comparison image 2
\includegraphics[width=0.95\textwidth]{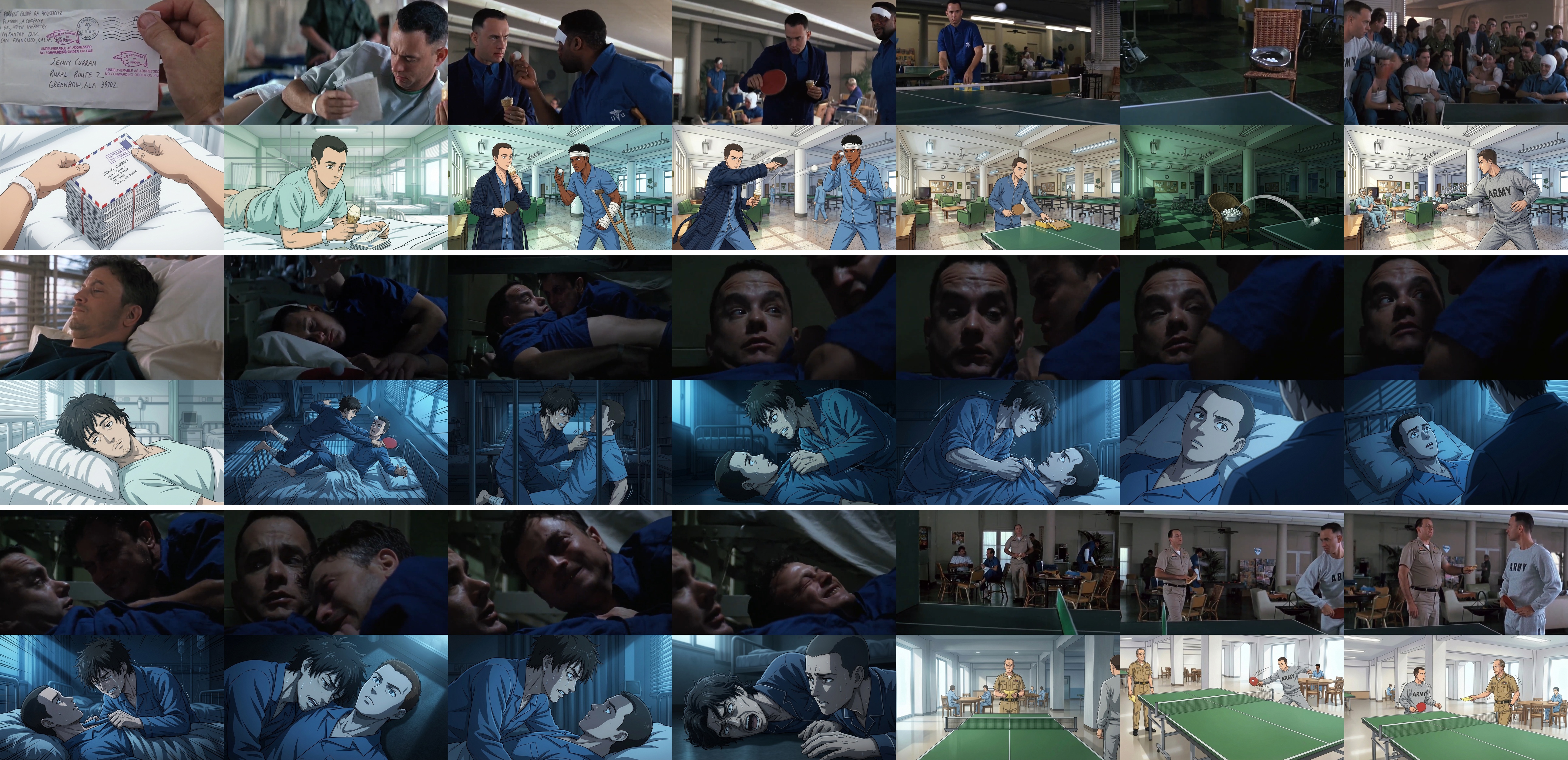}
\caption{\textbf{Forrest Gump (Style :} Anime)}
\label{fig:keyframe_result_2}
\end{figure}

\begin{figure}[h]
\centering
% TODO: Replace with actual keyframe comparison image 3
\includegraphics[width=0.95\textwidth]{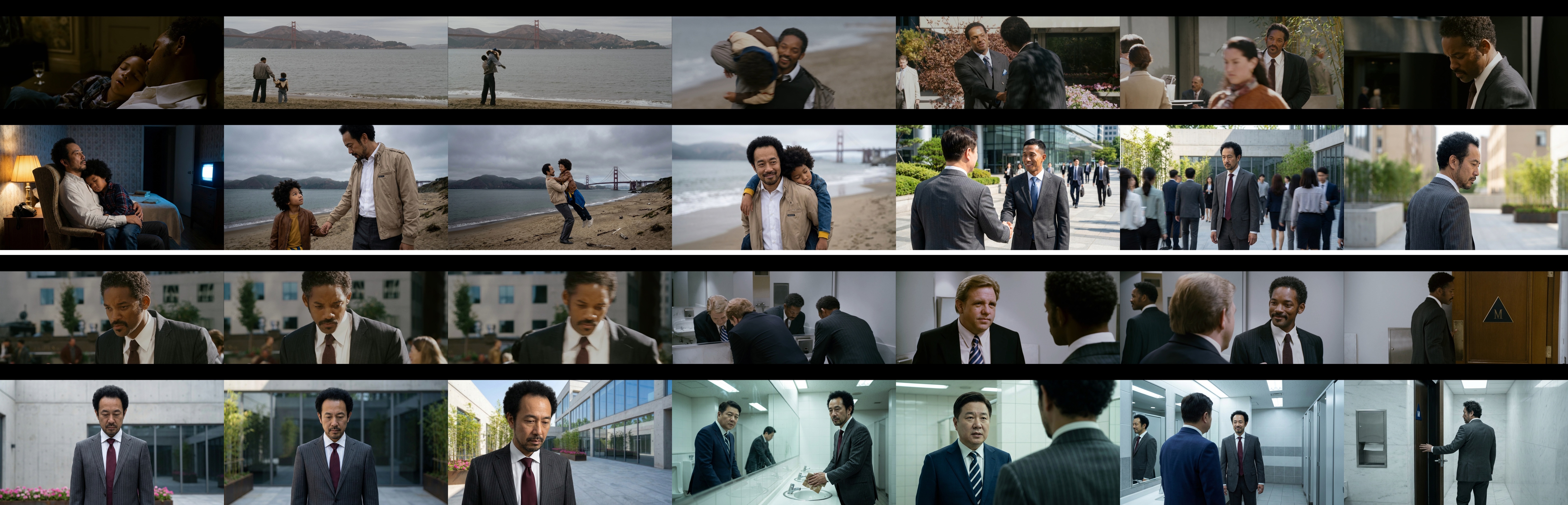}
\caption{\textbf{The Pursuit of Happiness (Style :} Realistic)}
\label{fig:keyframe_result_3}
\end{figure}

\begin{figure}[h]
\centering
% TODO: Replace with actual keyframe comparison image 4
\includegraphics[width=0.95\textwidth]{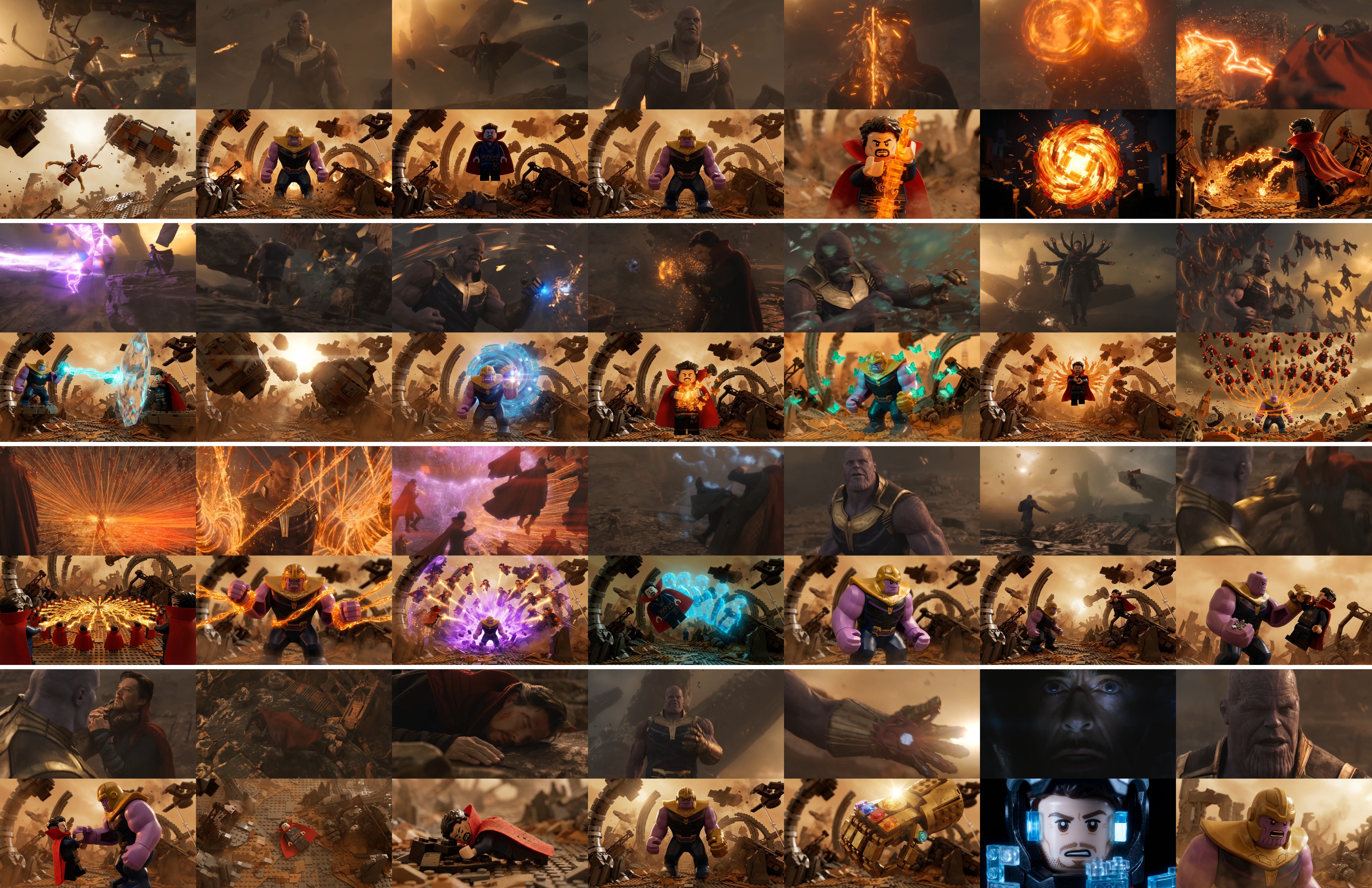}
\caption{\textbf{Avengers : Infinity War (Style :} Lego)}
\label{fig:keyframe_result_4}
\end{figure}

\begin{figure}[h]
\centering
% TODO: Replace with actual keyframe comparison image 5
\includegraphics[width=0.95\textwidth]{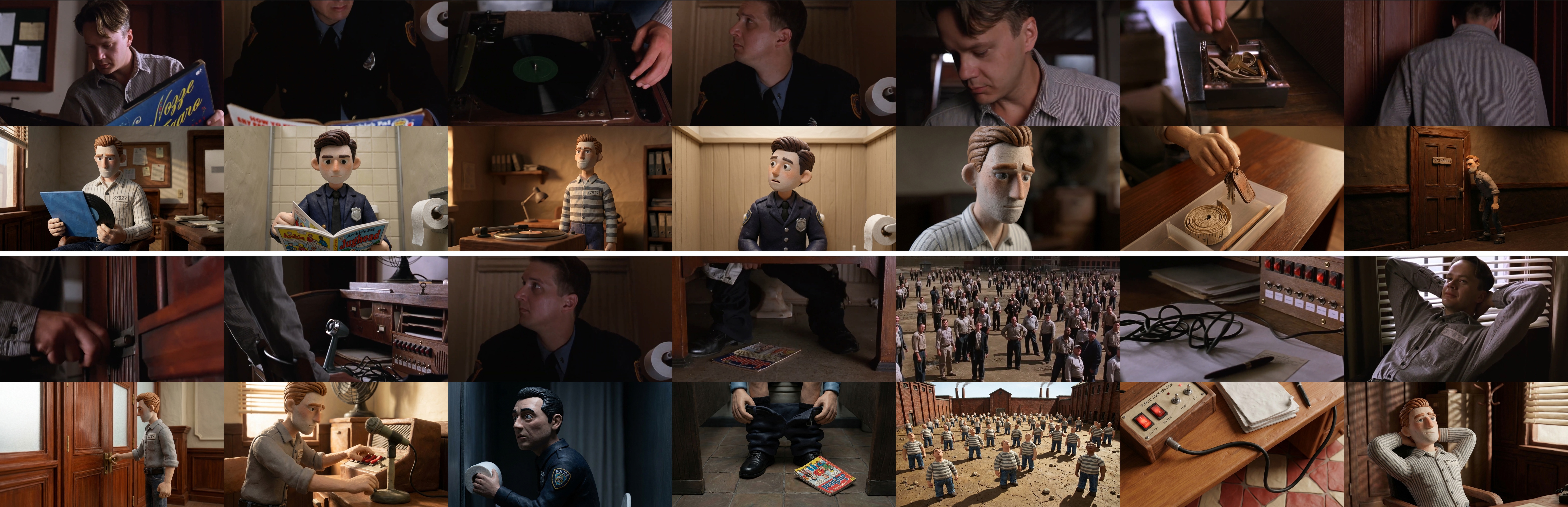}
\caption{\textbf{The Shawshank Redemption (Style :} Clay)}
\label{fig:keyframe_result_5}
\end{figure}

\begin{figure}[h]
\centering
% TODO: Replace with actual keyframe comparison image 6
\includegraphics[width=0.95\textwidth]{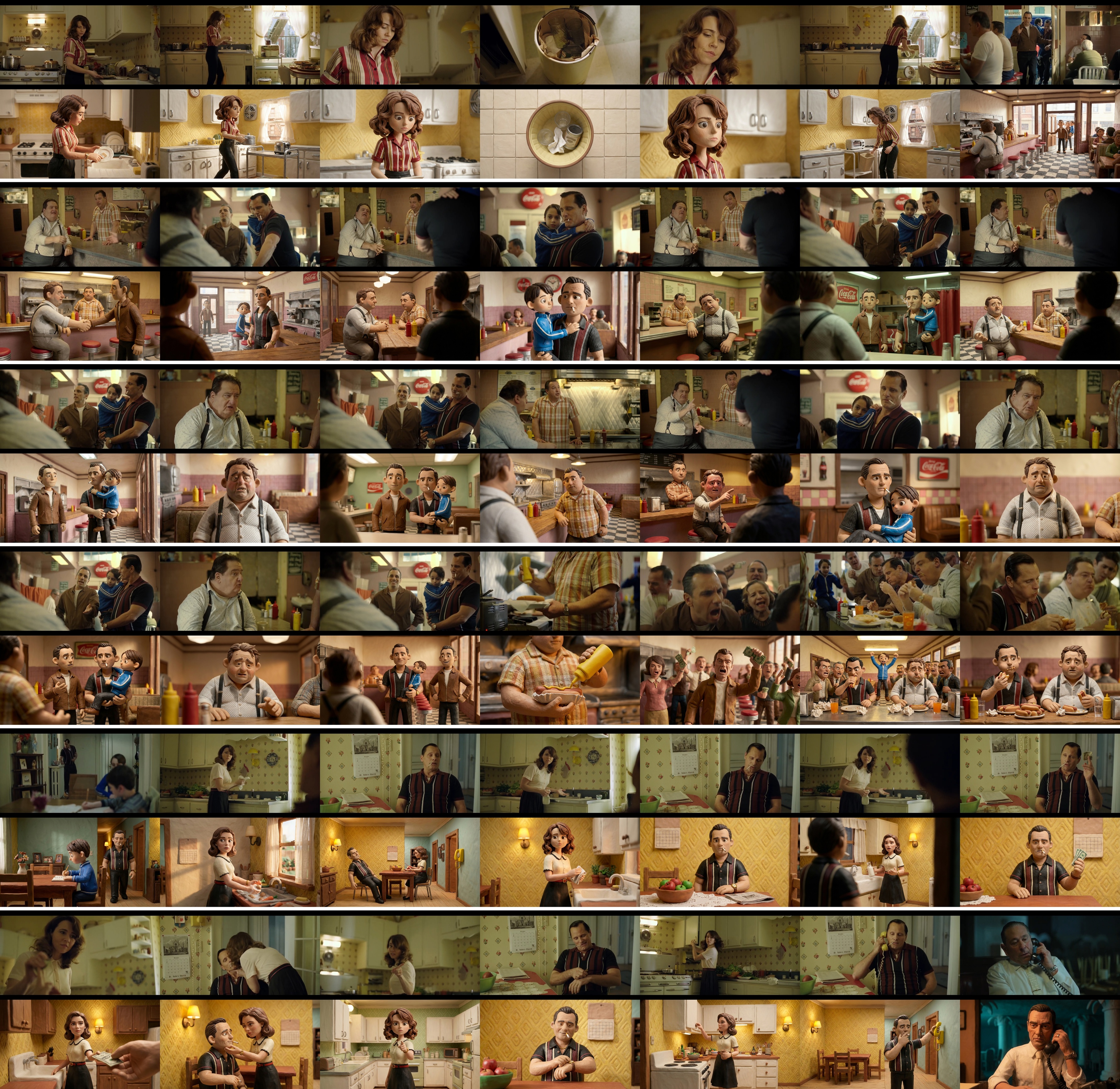}
\caption{\textbf{Green book (Style :} Clay)}
\label{fig:keyframe_result_6}
\end{figure}

\begin{figure}[h]
\centering
% TODO: Replace with actual keyframe comparison image 7
\includegraphics[width=0.95\textwidth]{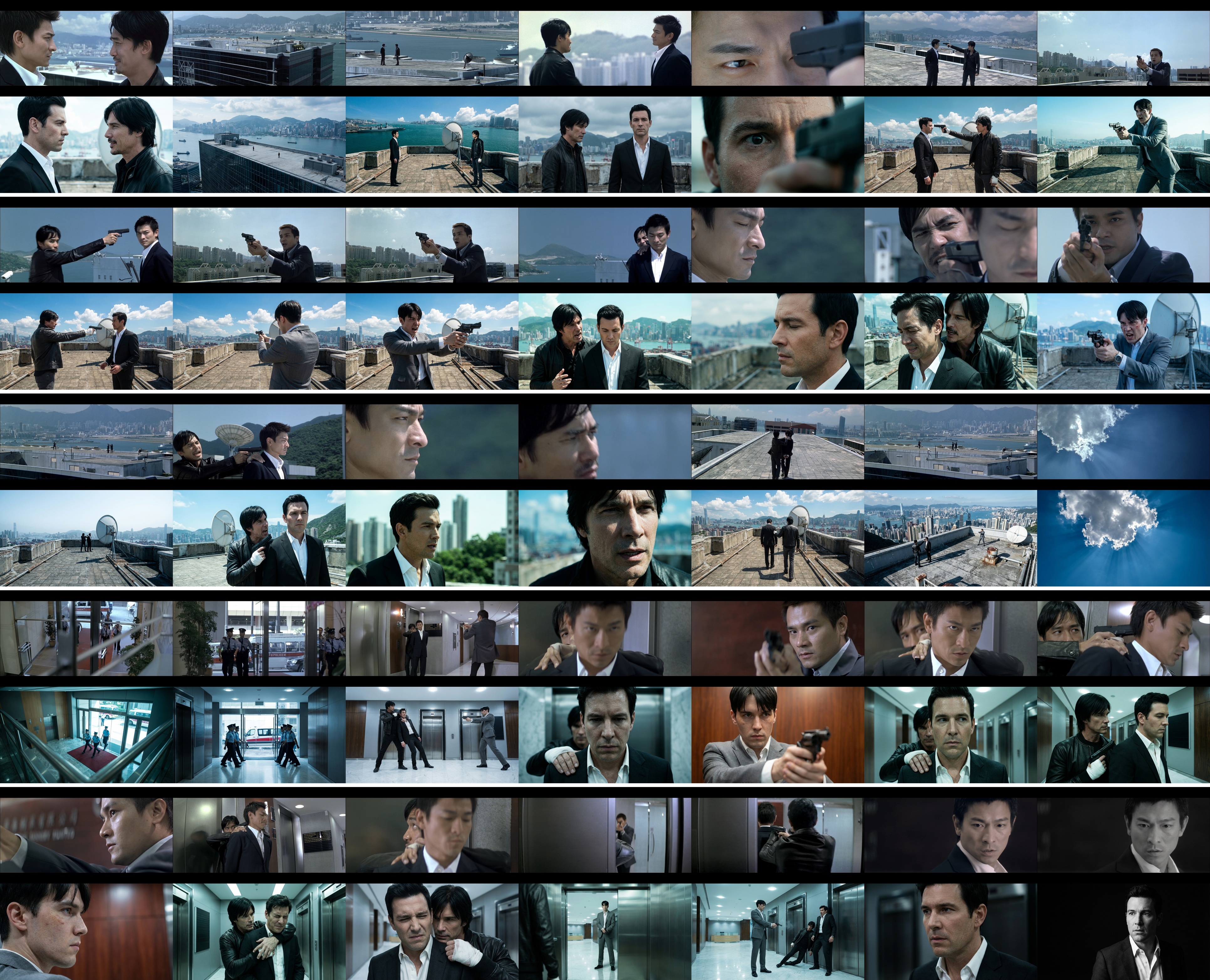}
\caption{\textbf{Infernal Affair (Style :} Realistic)}
\label{fig:keyframe_result_7}
\end{figure}

Additional keyframe-by-keyframe comparisons across our benchmark test videos are presented in Figures~\ref{fig:keyframe_result_1} through~\ref{fig:keyframe_result_7}, demonstrating the visual fidelity and consistency of our generation results against original video frames.

% \clearpage
% \newpage
% \input{checklist.tex}

\end{document}